\documentclass[]{elsarticle}

\usepackage{lineno,hyperref}
\usepackage{xspace}
\usepackage{enumerate}
\usepackage{amssymb}
\usepackage{amsmath}
\usepackage{graphicx}
\usepackage{relsize}
\usepackage{color,colortbl}
\usepackage{csquotes}
\usepackage{amssymb}
\usepackage{graphicx}
\usepackage{xspace}
\usepackage{dsfont}
\usepackage{textcomp}
\usepackage{url}
\usepackage{array}
\usepackage{pdflscape}
\usepackage{rotating}
\usepackage{blindtext}
\usepackage{longtable}
\usepackage{algpseudocode}
\usepackage[ruled,vlined]{algorithm2e}
\usepackage{caption}
\usepackage{appendix}
\usepackage{afterpage}

\newtheorem{Definition}{{\bf  Definition}}%[section]
\newtheorem{Lemma}{{\bf  Lemma}}%[section]
\newtheorem{Property}{{\bf  Property}}%[section]
\newtheorem{Proposition}{{\bf  Proposition}}%[section]
\newtheorem{Example}{{\bf  Example}}%[section]
\newcommand{\drule}[2]{\mbox{\ensuremath{#1 \at #2}}}
\newcommand{\at}{\ensuremath{\,-\!\!\!\prec\,}}
\newcommand{\LLa}{\ensuremath{{\mathcal L}}}

\newcommand{\ie}{\textnormal{\emph{i.e.}, }}

\newcommand{\facto}[1]{\ensuremath{#1}}
\newcommand{\argdrule}[2]{\drule{#1}{#2}}
\newcommand{\AForm}{\mbox{\ensuremath{\mathsf{A} = \langle \Slabel,\le, \OR,\OA, \OC,\IES, \IE \rangle}}}

\newcommand{\ARGPHI}{\ensuremath{\mathit{Arg}_\Phi}}

\newcommand{\Labelplus}[2]{\mbox{\ensuremath{\mu_{#1}^{#2}}}}
\newcommand{\Labelminus}[2]{\mbox{\ensuremath{\delta_{#1}^{#2}}}}

\newcommand{\nega}[1]{{\ensuremath{\sim}\xspace}#1}
\newcommand{\negaBar}[1]{\ensuremath{\overline{#1}}}

\newcommand{\Sup}{\ensuremath{\odot}}
\newcommand{\Con}{\ensuremath{\ominus}}
\newcommand{\Agr}{\ensuremath{\oplus}}

\newcommand{\support}{\ensuremath{\odot}}
\newcommand{\conflicto}{\ensuremath{\ominus}}
\newcommand{\agregation}{\ensuremath{\oplus}}
\newcommand{\neutro}{\ensuremath{\bot}}
\newcommand{\neutros}{\ensuremath{\top}}
\newcommand{\algebra}[1]{\ensuremath{\mathbf{#1}}}
\newcommand{\STR}[1]{\ensuremath{\mathbb{#1}}}
\newcommand{\Labeling}[1]{\ensuremath{\mathcal{V}(#1)}}

\newcommand{\ALL}{\ensuremath{\mathcal A}}
\newcommand{\Slabel}{\ensuremath{A}}
\newcommand{\R}{\ensuremath{\mathcal R}}
\newcommand{\K}{\ensuremath{\mathcal K}}
\newcommand{\F}{\ensuremath{\mathcal F}}

\newcommand{\OR}{\ensuremath{\odot}}
\newcommand{\E}{A}
\newcommand{\OA}{\ensuremath{\oplus}}
\newcommand{\OC}{\ensuremath{\ominus}}
\newcommand{\IE}{\ensuremath{\bot}}
\newcommand{\IES}{\ensuremath{\top}}

\newcommand{\Graph}[1]{\ensuremath{G}}

\newcommand{\LGraph}[1]{\ensuremath{G_{\mathcal{A}}}}
\newcommand{\CLGraph}[1]{\ensuremath{G_{R}}}

\newcommand{\model}{\ensuremath{\mathfrak{m}(\Phi)}}

\newtheorem{Theorem}{\bf Theorem}

\newcommand{\LAFARG}{\mbox{\ensuremath{\langle {\mathcal L}, {\mathcal R},  {\mathcal K}, {\mathcal A}, {\mathcal F} \rangle}}}

\journal{Approximate Reasoning - https://doi.org/10.1016/j.ijar.2016.12.016}

\begin{document}

\begin{frontmatter}

    \title{An Approach to Characterize Graded Entailment of Arguments through a Label-based Framework}

    \author[label1,label3]{Maximiliano C.\ D.\ Bud\'an}
    \author[label1]{Gerardo I.\ Simari}
    \author[label1,label2]{\\ Ignacio Viglizzo}
    \author[label1]{Guillermo R.\ Simari}

    \address[label1]{Departamento de Ciencias e Ing.\ de la Computaci\'on, Universidad Nacional del Sur (UNS) \\
                 Instituto de Ciencias e Ing.\ de la Computaci\'on (ICIC UNS--CONICET) \\
                 San Andres 800, $($8000$)$ Bah\'{i}a Blanca, Argentina}
    \address[label2]{Departamento de Matem\'atica, Universidad Nacional del Sur (UNS) \\
    Av.\ Alem 1253, $($8000$)$ Bah\'{i}a Blanca, Argentina}
    \address[label3]{Departamento de Matem\'atica, Universidad Nacional de Santiago del Estero\\
                     Belgrano(s) 1912, $($4200$)$ Capital, Sgo.\ del Estero, Argentina}

\begin{abstract}
	Argumentation theory is a powerful paradigm that formalizes a type of commonsense reasoning that aims to simulate the human ability to resolve a specific problem in an intelligent manner. A classical argumentation process takes into account only the properties related to the intrinsic logical soundness of an argument in order to determine its acceptability status. However, these properties are not always the only ones that matter to establish the argument's acceptability---there exist other qualities, such as strength, weight, social votes, trust degree, relevance level, and certainty degree, among others.
	
	In this work, we redefine the argumentative process to improve the analysis of arguments by considering their special features in order to obtain more refined results. Towards this end, we propose adding meta-level information to the arguments in the form of labels representing quantifiable data ranking over a range of fuzzy valuations. These labels are propagated through an argumentative graph according to the relations of support, conflict, and aggregation between arguments. Through this process we obtain final labels that are useful in determining argument acceptability.
\end{abstract} 

\end{frontmatter}

\section{Introduction}\label{Sec.Intro}

Artificial intelligence explores the implementation of systems that show an intelligent behavior when determining a solution to real-world problems. Such problems are usually immersed in contexts where the knowledge that describes the domain of discourse is incomplete or inconsistent.  Within the artificial intelligence field, argumentation is a human-like reasoning process that follows a commonsense strategy to resolve disagreements. In a general sense, argumentation can be defined as the study of the interaction among arguments for and against conclusions, with the purpose of determining which conclusions are acceptable in order to use such claims to resolve real-world problems. Argumentation tools are applied in many areas such as legal reasoning~\cite{FreemanFarley96,bench2008legal}, intelligent web search~\cite{chesnevar04,rahwan2007laying}, recommender systems~\cite{siersdorfer2009social,briguez2014argument}, autonomous agents and multi-agent systems~\cite{parsons02,2006:kais}, cyber-security~\cite{ShakarianSMP15}, and many others~\cite{Simari89,Carbogim00,2005:ccia}.

In traditional argumentation systems there is no notion of domain-dependent features associated with arguments or attacks. However, in many real-world problems it is necessary to provide further details representing their features in order to obtain more refined results. Such features may not only be based on properties of intrinsic logical soundness, but also other qualities can be taken into account to improve the acceptability process, seeking to obtain more information about argument acceptability. For instance, each argument may have associated different features, such as its strength~\cite{VBAF-BC}, weight~\cite{dunne2011weighted}, temporal availability intervals~\cite{MannHunterComma}, reliability varying with time~\cite{budan2014modeling}, among others. Continuing with the same intuition, other works have applied fuzzy theories~\cite{krause1995logic,stranders2008fuzzy,leite2011social} to enrich the expressive power of the classical argumentation model in two ways: represent the relative strength of the attack and support relationships between arguments, and represent the acceptability degree of arguments.

Based on these intuitions, the argumentation process based on valuations is defined in three steps: determine which are the domain-dependent attributes that will be associated with arguments, characterize the defined arguments to represent the domain of discourse, and determine the arguments' acceptability. In the first step, it is necessary to perform an analysis of the application domain to determine the features associated with the knowledge that describes such domain; for instance, reliability of the knowledge sources, user preferences, social votes, among others. In the second step, the attributes associated with arguments can be determined independently of the interactions with other arguments, as done in~\cite{VBAF-BC,leite2011social}, or those that are dependent on the relations (support and attack) that the argument has with other arguments~\cite{cayrol2005graduality,dunne2011weighted}. In the third step, it is possible to analyze the argument's acceptability in two ways: individual acceptability, where the acceptability of an argument depends on its attributes~\cite{VBAF-BC,krause1995logic}, and collective acceptability, where a set of arguments satisfies certain properties~\cite{dung1995acceptability,cayrol2005acceptability}. Recently, a combination of both points of view was considered, providing more information regarding argument acceptability and allowing to define a more gradual acceptability status~\cite{cayrol2005graduality,dunne2011weighted}.

In this work, we will develop a formalism that combines the power of these proposals by generalizing and providing a flexible structure to represent a specific domain with the objective of satisfying a particular goal. This formalization, called \emph{Labeled Argumentation Frameworks} (LAF), uses the knowledge representation capabilities provided by the \textit{Argument Interchange Format} (AIF)~\cite{AIF-GR} and leverages an \emph{Algebra of Argumentation Labels}, which allows to represent the arguments' features through labels, propagating them in the argumentation domain through a series of operations defined for this purpose. These argument labels are defined over the $[0,1]$ interval and describe the quality of arguments that can be affected by the interactions between arguments, such as their support, conflict, and aggregation. Once the propagation process is completed and the final argumentation labels associated with the arguments are obtained, we use this information for different purposes: to establish the arguments' acceptability according to different acceptance levels or acceptability classes, to specify when an argument is better than another (in order to resolve the conflicts caused by the context restrictions that should not be treated specifically as opposing reasons), to define an \emph{acceptability threshold} in order to determine whether an argument satisfies certain conditions to be accepted in a particular domain, and to analyze the possible models for a particular real-wold problem that determine the optimal justification for a certain conclusion, among others.

As was mentioned before, we will increase the representation capability of argumentation systems in order to represent the real-world features of the arguments through the use of labels, bringing the possibility of operating over these labels in the argumentative process. In this way, we contribute to the successful integration of argumentation in different artificial intelligence applications, such as autonomous agents in decision Support systems, knowledge management, recommender systems, and others of similar importance. Now, we will illustrate the usefulness of our proposal for a particular agent decision-making problem, where the reasons supporting decisions have associated degrees of trust and agent preferences.

\medskip
\textit{Consider the following scenario, in which Brian must decide whether to buy $\tt{house \ A}$ or $\tt{house \ B}$. To reach a decision, Brian ponders the following arguments provided by different sources}:

\medskip
\noindent
-- \textit{Carol, a real estate agent, advances the following arguments in favor of buying $\tt{house \ A}$ and against buying $\tt{house \ B}$}:

\begin{itemize}
	\item[] \textit{$\tt{house \ A}$ is located in a good area to live that has all the basic services. In addition, most of the neighbors are peaceful people, giving reasons to think that $\tt{house \ A}$'s area is quiet.}
	
	\item[] \textit{$\tt{house \ B}$ is located in a bad area to live, because it has precarious services, there is no public transportation nor education services nearby, increasing the cost of living for the residents. Also, $\tt{house \ B}$ has a high property tax debt.}	
\end{itemize}

\noindent
-- \textit{Daniel, also a real estate agent, poses the following arguments in favor of buying $\tt{house \ B}$ and against buying $\tt{house \ A}$}:

\begin{itemize}
	\item[] \textit{$\tt{house \ B}$ is located in a good area to live, since it is safe. %The $\tt{house \ B}$'s area has a low crime level.
		In addition, $\tt{house \ B}$'s area has a low level of environmental pollution, so it is a good area to live in.}
	
	\item[] \textit{$\tt{house \ A}$ has a high repair cost, since it has electrical and roof problems.}
\end{itemize}

\noindent
-- \textit{Ana, an architect, presents the following arguments in favor and against buying these houses}:

\begin{itemize}
	\item[] \textit{$\tt{house \ A}$ has a good orientation with regard to adequate natural lighting and the ventilation of its rooms.}
	
	\item[] \textit{$\tt{house \ B}$ has a good construction, since it was built with first-quality materials. Also, this house has a solid foundation and an adequate footing.}		
\end{itemize}

\noindent
-- \textit{\emph{Metro News}, a local newspaper, publishes an article about security in  $\tt{house \ A}$'s area where the following argument appears}:

\begin{itemize}
	\item[] \textit{The police force was reinforced last month, and the area is now safe. Given this development, it has not been chosen by gangs to operate.}
\end{itemize}

\noindent
-- \textit{\emph{Neighbors}; some rumors about the areas are represented in the following arguments}:
\begin{itemize}
	\item[] \textit{A gang was operating in $\tt{house \ A}$'s area, giving reasons to think that it is unsafe; therefore, this is not a good area to live in.}
	
	\item[] \textit{In $\tt{house \ B}$'s area, the police force was reinforced last month, increasing safety.}
\end{itemize}

\textit{Furthermore, Brian considers Ana to be more trustworthy than Carol, and Carol more trustworthy than Daniel. Also, he believes Daniel more than the newspaper, and considers that neighbors are at most as believable as Daniel. In addition, his priorities sorted from highest to lowest are: safety, services, neighbors, pollution level, tax debt, quality and problems with construction, house size, and house orientation. Note that Brian has enough money to buy only one house, so he must decide between $\tt{house \ A}$ and $\tt{house \ B}$ taking into account their characteristics.}

\medskip
The arguments have associated features that influence the final decision. In this way, the agent uses this information to determine which of these options is the most convenient. In our particular example, it would be interesting to assign to the arguments a trust and preference measurement, representing the trust that the agent gives to the information source and the preference of the agent over the characteristics of the houses, respectively. The proposed formalism allows the representation of argument features, propagating and combining these features in the argumentation domain interpreting the support, aggregation, and conflict relation among arguments---in particular, the trust degree of an acceptability status taking into account the preferences of the agent. In addition, our formalism specifies when an argument is better than another, in order to resolve the conflicts caused by the context constraints (for instance, ``Brian only has enough money to buy one house'').
Another important feature of our approach is that it automatically handles problems that could arise from cycles in the attack relation; this is done via the formulation of a system of equations that characterizes solutions to the overall assignments. Problems due to odd-length cycles have been identified in the literature for some time now and solutions proposed 
for abstract frameworks~\cite{baroni2003solving}
and possibilistic DeLP~\cite{alsinet2011maximal}.

This paper is structured as follows: in Section~2 we give a brief introduction to the Argument Interchange Format (AIF), which contains the elements we need for our development;
in Section~3 we introduce a particular abstract algebra for handling the labels associated with the arguments that we call {\em Algebra of Argumentation Labels}; the core contribution of the paper is presented in Section~4 as the formalism characterizing {\em Labeled Argumentation Frameworks} (LAF) together with an example of an application in the agent decision-making domain; in Section~5 we discuss related work, and in Section~6 we conclude and propose future work.

\section{Argument Interchange Format}\label{Sec.AIF}

\begin{figure}[t]
	\centering
	\includegraphics[width=1\textwidth]{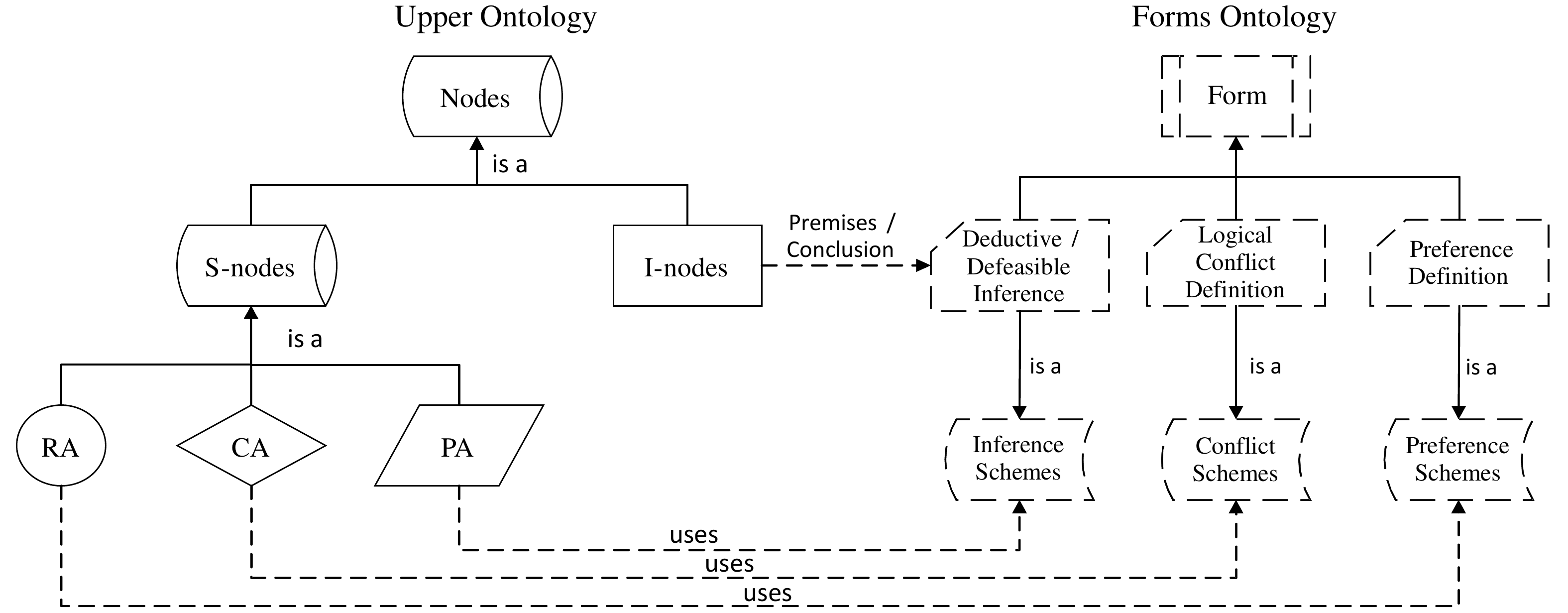}
	\caption{\small AIF Core Ontology}\label{Fig.AIFcore}
\end{figure}

The \emph{Argument Interchange Format} (AIF) is an ontology created to capture, in an abstract way, the specification of argumentative information that describes a discourse domain and the relations among these pieces of information, such as inference, conflict and preference. In AIF, arguments and their mutual relations are represented through a \textit{Typed Directed Graph}, which is an intuitive way to model arguments in a structured and systematic way without the formal constraints of a logic~\cite{AIF-GR}. The AIF core ontology is naturally defined in two parts: the \textit{Upper Ontology} and the \textit{Forms Ontology} (Figure~\ref{Fig.AIFcore}).

In the ontology, arguments and the relations between them are conceived of as an argument graph. The Upper Ontology defines the basic building blocks of AIF argument graphs, nodes, and edges. The Forms Ontology allows us to type the elements of the upper ontology in terms of argumentation-theoretical concepts. In AIF, there exist \textit{information nodes} (I-nodes) and \textit{scheme application nodes} (S-nodes), where I-nodes are used to represent propositional information contained in an argument (claim, premise, data, among others) and S-nodes capture the application of schemes, such as inference rule schemes (RA-nodes), conflict schemes (CA-nodes) and preference schemes (PA-nodes)\label{page:PA-node}.
S-nodes can be further classified; for example, inference schemes can be deductive or defeasible, and defeasible inference schemes can be subdivided into more specific argumentation schemes (e.g., Expert Opinion or Witness Testimony, among others). Thus, inference schemes, conflict schemes, and preference schemes in the Forms Ontology embody the general
principles expressing how it is that $q$ is inferable from $p$, $p$ is in conflict with $q$, and $p$ is preferable to $q$, respectively. The individual RA-nodes, CA-nodes, and PA-nodes that fulfill these schemes then capture the passage or the process of actually inferring $q$ from $p$, conflicting $p$ with $q$ and preferring $p$ to $q$, respectively.

This collection of nodes is used to build a typed directed graph, called \emph{Argument Network}, which is defined as follows:

\begin{Definition}[Argument Network]\label{Def.argumentnetworkAIF}
	An AIF argument network is a digraph $G = (V, E)$, where:
	\begin{enumerate}
		\item[--] $V = I \cup \mathit{RA} \cup \mathit{CA} \cup \mathit{PA}$ is the set of nodes in $G$, where $I$ are the I-nodes, $RA$ are the RA-nodes, $CA$ are the CA-nodes, and $PA$ are the PA-nodes;
		\item[--] $E \subseteq V \times V \setminus I \times I$ is the set of edges in $G$. Any edge $e \in E$ is assumed to have exactly
		one type among the following: premise, conclusion, preferred element, dispreferred element, conflicting element, conflicted element;
		\item[--] if $v \in \mathit{RA}$ then $v$ has at least one direct predecessor via a premise edge and exactly one direct successor via a conclusion edge;
		\item[--] if $v \in \mathit{PA}$ then $v$ has exactly one direct predecessor via a preferred element edge and exactly one direct successor via a dispreferred element edge;
		\item[--] if $v \in \mathit{CA}$ then $v$ has exactly one direct predecessor via a conflicting element edge and exactly one direct successor via a conflicted element edge.
	\end{enumerate}
\end{Definition}

Basically, there are two types of edges: \emph{scheme edges} emanate from S-nodes and connected with either I-nodes or S-nodes, and \emph{data edges} emanating from I-nodes but necessarily ending in S-nodes. Notice that edges connecting I-nodes are forbidden, because I-nodes cannot be connected without an explanation that justifies that connection. There is always a scheme, justification, inference or rationale behind a relation between two or more I-nodes that is captured through an S-node. Moreover, only I-nodes can have zero incoming edges, as all S-nodes relate two or more components (for RA-nodes, at least one antecedent is used to support at least one conclusion; for PA-nodes, at least
one alternative is preferred to at least one other; and for CA-nodes, at least one claim is in conflict with at least one other).

In this work we will concentrate only on the relationships that are relevant to our formalization, which are: I-to-RA and RA-to-I representing the premises and the conclusion of an inference scheme application, respectively; and I-to-CA and CA-to-I representing the relation between conflicting or contradictory premises or claims--PA-nodes will arise later on (cf.\ Page~\pageref{page:preference}) as a byproduct of the analysis of the labels computed for each of the I-nodes.
As we said before, inference schemes in the AIF ontology are similar to the rules of inference in a logic, in that they express the general principles that form the basis for actual inference. They can be deductive (e.g., the inference rules of propositional logic~\cite{reichenbach1947elements}) or defeasible (e.g., argumentation schemes~\cite{walton2013argumentation}).

Given a network, it is possible to identify arguments; a simple argument can be represented by linking a set of I-Nodes denoting premises to an I-Node denoting a conclusion via a particular RA-Node. Formally, we have:

\begin{Definition}[Simple Argument]\label{Def.simpleargumentAIF}
	Let $G = (V, E)$ be an AIF argument network with $V = I \cup RA \cup CA \cup PA$. A simple argument in $G$ is a tuple $(P, R, C)$ where $P \subseteq I, C \in I$, and $R \in RA$,
	such that for all $p \in P$ there exists $(p, R) \in E$ and $(R, C) \in E$.
\end{Definition}

\begin{figure}[t]
	\begin{center}
		\includegraphics[width=0.6\textwidth]{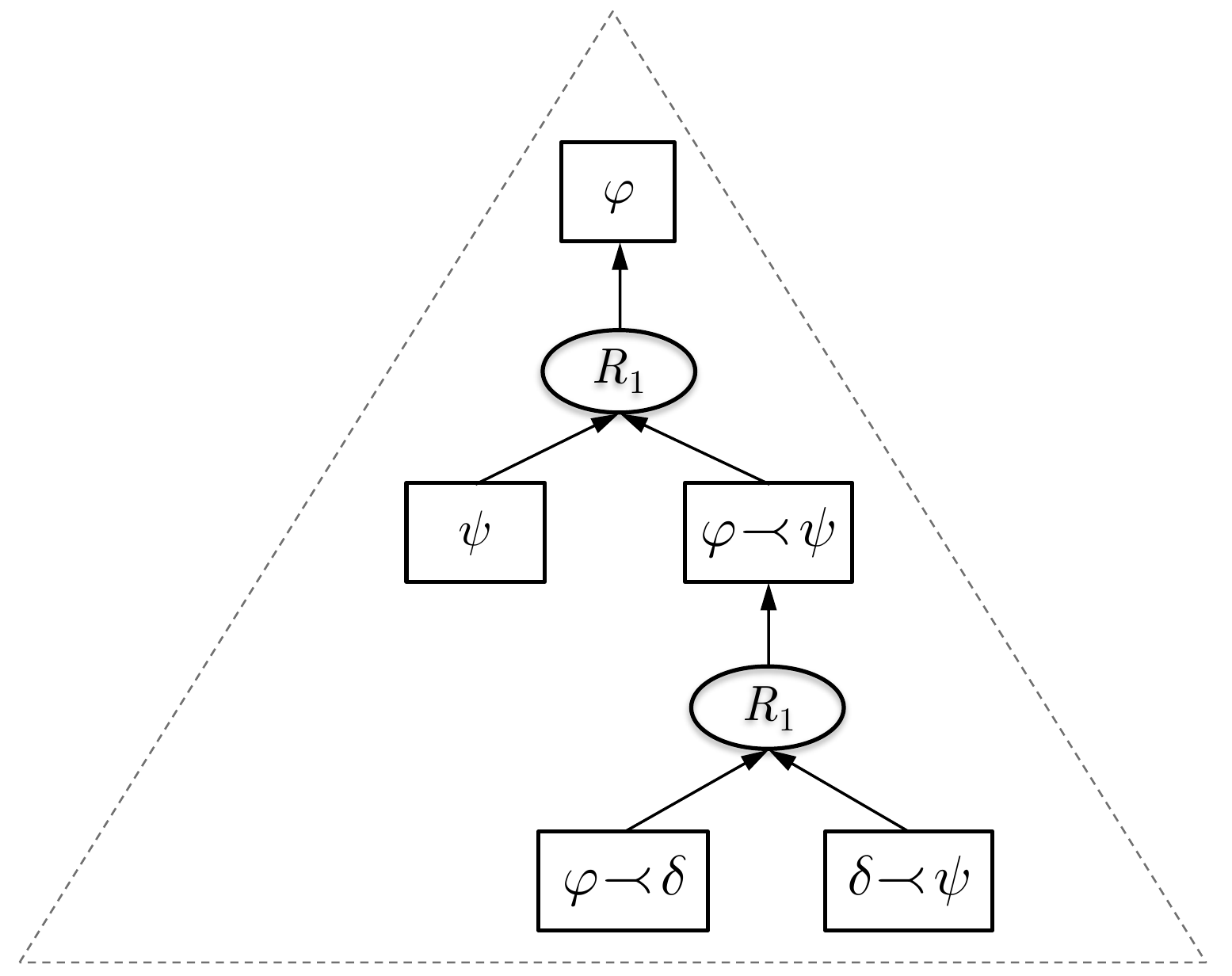}
		\caption{Representation of an argument in AIF}\label{Fig.argument}
	\end{center}
\end{figure}

In AIF, a simple argument $\mathtt{A}$ supports an argument $\mathtt{B}$ if the conclusion of $\mathtt{A}$ is part of the premises that support the conclusion of $\mathtt{B}$. Then, simple arguments can be aggregated into a more complex argument structures called arguments.
In Figure~\ref{Fig.argument}, we present an example of a complex argument structure where there exist two inference scheme classes combined to support a particular conclusion: the inference scheme that instantiates RA-node $R_1$ is Defeasible Modus Ponens with premises $\psi$ and \drule{\varphi}{\psi} (here, \drule{}{} is a connective standing for defeasible implication), and the conclusion is $\varphi$; the inference scheme that instantiates RA-node $R_2$ is Defeasible Hypothetical Syllogism with premises \drule{\psi}{\delta} and \drule{\delta}{\varphi} and conclusion
\drule{\psi}{\varphi}. In both cases, $\varphi$ and $\psi$ are meta-variables ranging over well-formed formulas in some language.

Furthermore, it is possible that two or more arguments share the same conclusion. This corresponds to the notion of {\em argument accrual} developed in [33, 28, 25], where the strength of the shared conclusion is the aggregation of the strengths of each individual argument supporting it.

\begin{Definition}[Argument Accrual]\label{Def.accrualargumentAIF}
	Let $G = (V, E)$ be an AIF argument network with $V = I \cup RA \cup CA \cup PA$, \ARGPHI\ be a set of arguments, and $\ARGPHI^Q$ be a set of arguments supporting the same conclusion $Q$, where $\ARGPHI^Q \subseteq \ARGPHI$. We define the argument accrual for $Q$ from the set $\ARGPHI^Q$ as the union of these arguments (with a single, shared root).
\end{Definition}

\begin{figure}[t]
	\begin{center}
		\includegraphics[width=1\textwidth]{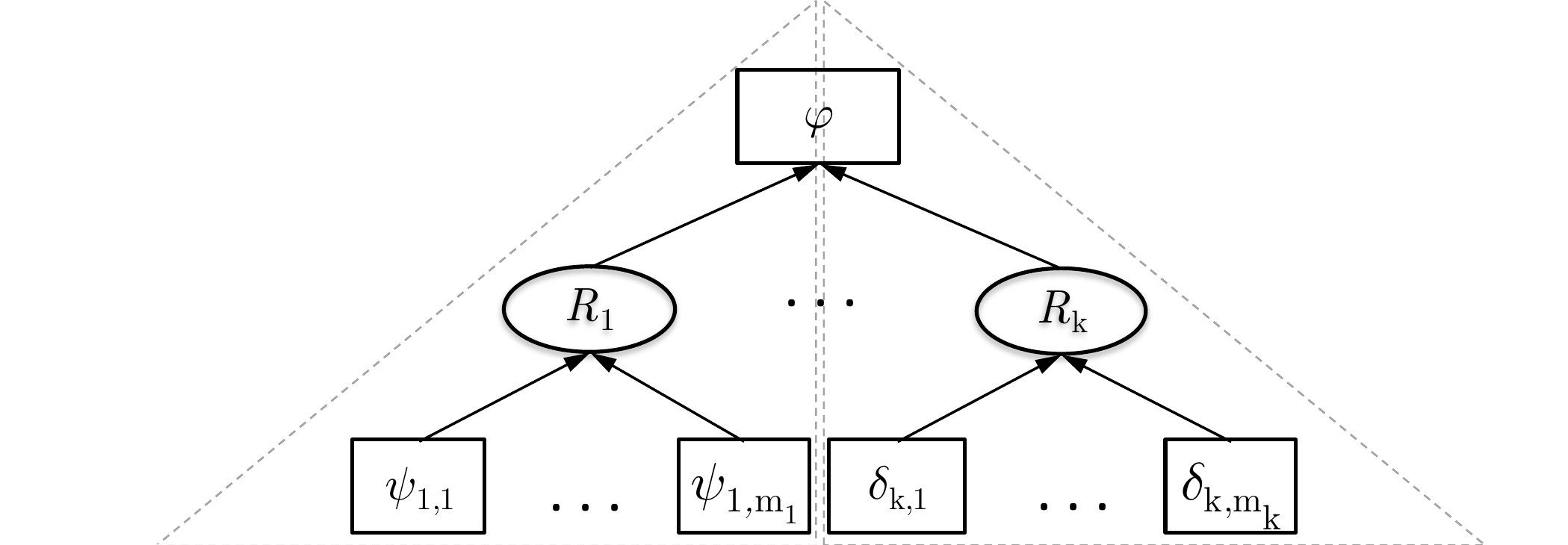}
		\caption{Representation of an accrual argument in AIF}\label{Fig.argumentaccrual}
	\end{center}
\end{figure}

In Figure~\ref{Fig.argumentaccrual}, we present an example of an accrual argument where there exist $k$ simple arguments supporting the conclusion $\varphi$ through their corresponding set of premises that justify such conclusion. Note that each simple argument has a particular RA-node that can be instantiated with a specific inference scheme.

The abstract AIF ontology as presented here is purely intended as a language for expressing arguments. In order to do anything meaningful with such arguments (e.g., visualize, query, evaluate, and so on), they must be expressed in a more concrete language so that they can be processed by additional tools and methods. For instance, the components of an argument network can be instantiated using propositional logic, as is demonstrated by Bex \emph{et al.}\ in~\cite{bex2012logical}, where one of the aims is to show how AIF argument graphs can be evaluated; that is, how a certain defeat status can be assigned to the elements of an argument graph using the classical argumentation-theoretic semantics. Another example of this instantiation is shown in~\cite{rahwan2007laying}, where the authors refined the abstract ontology in RDF, a Semantic Web-based ontology language, which may then be used as input for a variety of Semantic Web argument annotation tools. In a similar way,
Rahwan~\cite{rahwan2008mass} has formalized AIF in Description Logics, which allows for the automatic classification of schemes and arguments. In this same vein, in Section~4 we present a formalism that allows to represent the knowledge that describes a specific domain through a concrete argumentation network. Then, we analyze the acceptability status of such knowledge taking into account their features that describe the quality of knowledge, which can be affected by their interactions in the argumentation domain. 

\section{Algebra of Argumentation Labels}

The real world is uncertain and vague---this is, in general terms, one of the main reasons why reality cannot be studied in absolute terms with techniques applicable only to certain situations. Thus, in order to obtain accurate models, real-world situations are adjusted to rigid and static mathematical models, which causes a loss of valuable information. In this context, Zadeh~\cite{zadeh2014fuzzy} argues that, in dealing with the study of high complexity phenomena, our ability to perform precise and significant formulations about their behavior diminishes until it reaches a threshold where the accuracy and relevance turn into mutually exclusive characteristics. Based on this intuition, argumentative formalisms should possess the ability to represent not only the logical soundness of arguments, but also other
domain-dependent features to provide a representation that is better adjusted to reality.

In this section, we introduce the use of labels as a tool for aiding in the assessment of arguments, which helps in their characterization and evaluation. To be useful, these labels must represent information about the arguments and how the arguments interact. A natural way of representing this information is to use a scale that measures a particular feature of the argument, such as trust, user preferences, or information accuracy. We will consider fuzzy valuations ranging between two distinguished elements: $\bot$ and $\top$, where $\bot$ represents the least possible degree in which an argument may possess a certain attribute, and $\top$ the maximum. In his work, Zadeh provides a set of tools for dealing with fuzzy values in the real interval $[0,1]$ as if it were a set of truth values of propositions~\cite{zadeh65fuzzy}. We borrow some of those tools as explained below, and introduce a new operation. The effectiveness of these methods has been proven by many applications; for  example, cf.~\cite{dubois1985review,klir1997fuzzy,zimmermann01fuzzy}.

\subsection{Elements of an Algebra of Argumentation Labels}

We define an \emph{Algebra of Argumentation Labels} as an abstract algebraic structure that contains the operations related to argument manipulation in the domain. The effect of aggregation, support, and conflict of arguments will be reflected in their labels, informing how the arguments have been affected by their interaction. The algebra is based on an ordered set, allowing the comparison of labels; this set is also characterized in an abstract way to allow the adaptation to different applications.
\begin{Definition}[Algebra of Argumentation Labels]\label{Def.Algebra_Labels}
	An algebra of argumentation labels is a $6$-tuple \AForm\ where:
	\begin{itemize}
		\item[--] \E\ is a set of labels called the \emph{domain of labels}.
		\item[--] $\le$ is a partial order over \E\ $($that is, a reflexive, antisymmetric, and transitive relation$)$.
		\item[--] $\top$ and $\bot$ are two distinguished elements of \E. $\top$ is the last label  with respect to $\le$, while $\bot$ is the first.
		\item[--] $\OR: \E \times \E \rightarrow \E$ is called a \emph{support operation} and satisfies:
		\begin{itemize}
			\item[] $\OR$ is a \textit{commutative} operation: for all $\alpha, \beta \in \E, \ \alpha \OR \beta = \beta \OR \alpha$.
			\item[] $\OR$ is  \textit{monotone}: for all $\alpha, \beta, \gamma \in \E$, if $\alpha \le \beta$, then
			$\alpha \OR \gamma \le \beta \OR \gamma$.
			\item[] $\OR$ is  \textit{associative}: for all $\alpha, \beta, \gamma \in \E$, $\alpha \OR (\beta \OR \gamma) =
			(\alpha \OR \beta) \OR \gamma$.
			\item[] \IES\ is the  \textit{neutral element} for \OR: for all $\alpha \in \E, \ \alpha \OR \IES = \alpha$.
		\end{itemize}
		\item[--]  $\OA: \E \times \E \rightarrow \E$ is called an \emph{aggregation operation} and satisfies:
		\begin{itemize}
			\item[] $\OA$ is a \textit{commutative} operation: for all $\alpha, \beta \in \E, \ \alpha \OA \beta = \beta \OA \alpha$.
			\item[] $\OA$ is \textit{monotone}: for all $\alpha, \beta, \gamma \in \E$, if $\alpha \le \beta$,
			then $\alpha \OA \gamma \le \beta \OA \gamma$.
			\item[] $\OA$ is \textit{associative}: for all $\alpha, \beta, \gamma \in \E$, $\alpha \OA (\beta \OA \gamma) =
			(\alpha \OA \beta) \OA \gamma$.
			\item[] $\IE$ is the neutral element for \OA: for all $\alpha \in \E, \ \alpha \OA \IE = \alpha$.
		\end{itemize}
		\item[--] $\OC: \E \times \E \rightarrow \E$ is called a \emph{conflict operation} and satisfies:
		\begin{itemize}
			\item[] For all $\alpha, \beta \in \E$, $\alpha \conflicto \beta \le \alpha$ if $\beta < \alpha$.
			\item[] For all $\alpha, \beta \in \E, \ \alpha \conflicto \beta = \neutro$ if $\beta \geq \alpha$.
			\item[] $\OC$ is \textit{monotone}: for all $\alpha, \beta, \gamma \in \E$, if $ \alpha \le \beta$,
			then $\alpha \OC \gamma \le \beta \OC \gamma$.
			\item[] $\bot$ is the neutral element for \OC: for all $\alpha \in \E, \ \alpha \conflicto \neutro = \alpha$.
			\item[] For all $\alpha, \beta \in \E$, if $\alpha \conflicto \beta = \neutro$ and $\beta \conflicto \alpha = \neutro$, then $\alpha = \beta$.
			\item[] For all $\alpha, \beta \in \E$, if $(\alpha \agregation \beta) < \top$, then $\big((\alpha \agregation \beta) \conflicto \beta\big) = \alpha$.
		\end{itemize}
	\end{itemize}
	
\end{Definition}

\subsection*{Support Operator}

\noindent
The support operation, denoted with $\OR$, is used to determine the valuation of a single argument based on the valuations of the arguments that support it. It is clear that one wants this operation to be invariant with respect to the order in which the supporting arguments are considered, and therefore the operation is both commutative and associative, with $\top$ as the neutral element. Furthermore, if an argument is supported by stronger arguments, its valuation must be higher than that of one with lesser valuations for its supporters, giving place to the monotonicity property (which is justified by the weakest-link principle). These conditions can be summarized by saying that $\OR$ is a \emph{t-norm}~\cite{schweizer61asociative,schweizer63asociative}. Note that the use of t-norms to represent knowledge behavior is widely recognized in different application domains; for example, Dubois and Prade~\cite{dubois1982class} use t-norms to model the uncertainty associated with subjective evidence in the analysis of legal cases. In another vein, Lukasiewicz and Straccia~\cite{lukasiewicz2008managing} analyze the efficiency of t-norms to model the uncertainty and precision of information in the Semantic Web domain.

\subsection*{Aggregation Operator}

\noindent
Now, if we have several arguments that share the same conclusion,
the aggregation operation, denoted with $\OA$, determines the valuation of an accrued argument based on the valuations of such arguments (i.e., the various sources) that support the conclusion. Thus, the valuation associated with an argument that can be supported by independent sources can be obtained by aggregating the valuations of such sources. The most natural way of doing this would be to directly add these, considering the conditions imposed by the domain. This is why the $\OA$ operation has some of the properties of addition of real numbers---it is commutative and associative, with $\bot$ as neutral element (arguments with the least possible valuation do not affect the accrual valuation).
Furthermore, the $\OA$ operation has the monotonicity property, which ensures that the valuation of a conclusion does not decrease if the valuation of the arguments that support such conclusion increase. These conditions on $\OA$ may be summarized by saying that it is a \emph{t-conorm}.
These operators are used in different application domains; for example, Grabisch \emph{et al.}~\cite{dubois1982class} present different ways to perform aggregation of arguments based on user preference through t-conorms. Another example is the work of Krause \emph{et al.}~\cite{lukasiewicz2008managing}, which introduces a series of criteria to perform aggregation of arguments supporting a particular conclusion in decision-making support systems, taking into account the uncertainty level associated with these arguments. The authors highlight the use of t-conorms as a sensible way to obtain the uncertainty level of a conclusion supported by multiple arguments.

\subsection*{Conflict Operator}

The conflict operation, denoted with $\OC$, determines the valuation of an argument after considering the valuations associated with opposing reasons. The conditions on the $\OC$ operation state that it behaves with respect to the aggregation operation in a similar way as subtraction acts with respect to the addition of real numbers. We have not chosen, however, to make $\OC$ simply a residue of the accrual operation $\OA$ in order to give the conflict operator the properties we deemed necessary to ensure a consistent conflict resolution within the argumentation domain. Thus, the $\OC$ operation has the anti-monotonicity property, ensuring that the valuation of a conclusion decreases if the valuation of the reasons against such conclusion increase. Note that if an argument has a label with valuation $\alpha \OA \beta$ because it has accrued the valuations of other arguments, and then it is attacked by an argument with valuation $\beta$, and $\alpha \OA \beta < \top$, its valuation becomes reduced to $\alpha$.  In case that $\alpha \OA \beta = \top$, some information is lost, so $\OC$ is not strictly an inverse of the accrual. Furthermore, $\bot$ is the neutral element for $\OC$, specifying that the valuation associated with an argument is not affected by counterarguments with the least possible valuation.

\subsection{Objectives and Intuitions behind an Algebra of Argumentation Labels}

Next, we will show how labels composed of elements from our algebra allow to represent, combine, and propagate the characteristics associated with arguments, and use this information for different purposes:
(i) evaluate and establish the acceptability status associated with arguments. The information provided by labels shall cooperate in the acceptability process, capturing knowledge behavior in the argumentation model created to represent a particular situation. For example, if an argument has associated a reliability degree, we can consider that an argument is accepted if its reliability is greater than $\bot$ after considering the effects of its related arguments;
(ii) represent the quantitative and qualitative qualification of arguments. This extension allows the possibility of analyzing conflict resolution between arguments from a new perspective, where the qualities of an argument decrease in relation to the qualities of opposing reasons;
(iii) establish acceptability degrees by using the information captured in the argumentative labels. Thus, the acceptability status of an argument depends on its characteristics;
(iv) specify a (partial or total) preference order over the set of arguments. This relation will be useful for different purposes, such as solving conflicts between arguments;
(v) introduce a threshold to establish the minimal requirements that an argument must satisfy to justify a particular statement, decision, or action; and
(vi) analyze a set of models that represent the different final characteristics associated with arguments, with the intention of identifying the solution that optimizes the qualities of a set of conclusions.

There are many possible examples of algebras of argumentation labels, and it is important to determine the most appropriate one to use in each case. This is a methodological question involving the semantics of the domain that could be tackled by devising experiments using examples where the desired conclusion is well known, or by performing tests using the cognitive evaluation of human subjects to approximate their assessment of the valuations obtained after their interactions. We have chosen functions ranging over partially ordered sets with additional operations as our representation. A full discussion of the generality of this choice for representing uncertain information can be found in \cite{walley00unified}. 

\section{Labeled Argumentation Framework}\label{Section.LAF}

The main objective of an argumentative framework is to imitate the reasoning mechanism often applied by humans to solve problematic situations in an intelligent way, taking into account incomplete and contradictory information. In this kind of frameworks it is useful to attach additional information about special characteristics of the arguments. For example, arguments could be built from an agent's knowledge, where each of them has a reliability measure of the information source attached. In the end, the agent determines their action based on the most reliable information that they have.

In this section we will focus on the development of a formalism called \emph{Labeled Argumentation Framework} (LAF) that combines the knowledge representation features provided by AIF with the processing of meta information using the algebra of argumentation labels. This framework will allow us to represent arguments taking into account their internal structure, the interactions between arguments, and special features of the arguments through argumentation labels. The effects produced by the interactions of support, conflict, and aggregation among arguments are reflected by the operations defined in the algebra of argumentation labels. So, as mentioned above, the final label attached to each argument is obtained; the information carried by this label is useful to establish
the acceptability status of arguments with additional information,
when an argument is better than another,
establish different acceptability degrees based on the qualities of the arguments representing the veracity of claims in the argumentation domain,
define an acceptability threshold, and
provide the ability to analyze the possible argumentation models for a problematic situation that optimize the justification of a particular conclusion.

In~\cite{Budan2015LAF}, we present an earlier version of this formalism; in this work, we introduce the treatment of a different kind of conflict cycles involved in an argumentation graph. Towards this end, we derive an algorithm that creates a system of equations representing the constraints that all valuations associated with the knowledge of the argumentation graph must fulfill---our algorithm obtains such a system independently from  the kind of argumentation graph used. In addition, we consider the possibility of optimizing the attributes associated with specific knowledge provided by the user, and we also present some interesting results related to the computability of the arguments' acceptability.

\subsection{Elements of a Labeled Argumentation Framework}

In LAF, we use the AIF ontology as the underlying knowledge representation model
to represent the internal structure of the arguments and the relations that describe their behaviors in the argumentation domain. We now present the basic elements that compose LAF, and the different proposals to determine argument acceptability when inconsistencies arise.

\begin{Definition}[Labeled Argumentation Framework]\label{Def.LAF}
	A Labeled Argumentation Framework $($LAF$)$ is a 5-tuple of the form $\Phi = $ \LAFARG\ where:
	
	\begin{itemize}
		\item[--] \LLa\ is a logical language for knowledge representation $($claims$)$ about the domain of discourse. We assume that the connectives of this language include one distinguished symbol ``$\nega$" denoting strong negation.
		
		\item[--] \R\ is a set of inference rules $R_1, R_2,\ldots, R_n$ defined in terms of \LLa\ $($\ie with premises and conclusion in \LLa$)$.
		
		\item[--] \K\ is the knowledge base, a set of formulas of \LLa\ describing knowledge about the domain of discourse.
		
		\item[--] \ALL\ is a set of algebras of argumentation labels $($cf.\ Definition~\ref{Def.Algebra_Labels}$)$
		$\mathsf{A}_1, \mathsf{A}_2,\ldots,$ $\mathsf{A}_n$, one for each feature that will be represented by the labels.
		
		\item[--] \F\ is a function that assigns to each element of \K,  an $n$-tuple of elements$\,$\footnote{When no confusion can occur we will follow the usual convention of mentioning elements in an algebra instead of referring to elements in the corresponding carrier set of that algebra.} in the algebras $\mathsf{A}_i, i=1,\ldots, n$. This is,  $\F: \K \longrightarrow \mathsf{A}_1\times \mathsf{A}_2\times\ldots\times \mathsf{A}_n$. We will use $\F_i$ to denote the element $\mathsf{A}_i$ (so, the projection on the $i$-th element of $\F$'s image.
	\end{itemize}
\end{Definition}

We use language \LLa\ to specify the knowledge base, and the set of inferences is specified by inference rules representing patterns of reasoning such as deductive inference rules (\emph{modus ponens}, \emph{modus tollens}, \emph{etc.}), defeasible inference rules (\emph{defeasible modus ponens}, \emph{defeasible modus tollens}, \emph{etc.}), or argumentation schemes (\emph{expert opinion}, \emph{Position to Know}, \emph{etc.}), among others. Every formula $\nega\nega\varphi$ of \LLa\ is considered equivalent to $\varphi$. Thus, we can assume that no subexpression of the form ``$\nega\nega\varphi$'' appears in the formulas of the language, yet the set of formulas in  \LLa\ is closed with respect to ``$\nega$''. It is important to note that the use of two or more consecutive ``$\nega$" in $\mathcal L$ is not allowed in order to simplify the definition of conflict between claims---this does not limit its expressive power or generality of the representation. We denote with $\overline{\varphi}$ the negation of a formula in $\mathcal L$, so $\overline{\varphi}$ is $\nega\varphi$ and $\overline{\nega\varphi}$ is simply $\varphi$.

\begin{Example}\label{Example_InstantiateLAF}
	Consider the following Labeled Argumentation Framework $\Phi = \LAFARG$:
	
	\medskip
	\noindent
	$-$ \LLa\ is a language defined in terms of two disjoint sets: a set of presumptions and a set of defeasible rules, where a presumption is a ground atom $X$ or a negated ground atom $\nega X$, where $\nega$ represents strong negation; a defeasible rule is an ordered pair, denoted $\argdrule{\mathtt{C}}{\mathtt{P}_1, \ldots,  \mathtt{P}_n}$, whose first component $\mathtt{C}$ is a ground atom, called the conclusion and the second component $\mathtt{P}_1, \ldots ,\mathtt{P}_n$ is a finite non-empty set of ground atoms, called the premises.
	
	\medskip
	\noindent				
	$-$ \R = \{\,dMP\}, consisting of the following inference rule:
	
	\medskip
	\hspace*{16pt}dMP:\ {\large $\frac{\mathtt{P}_1, \ldots, \mathtt{P}_n     \     \     \   \mathtt{C} {\scriptstyle\at} \mathtt{P}_1, \ldots,  \mathtt{P}_n} {\mathtt{C}}$} $($Defeasible Modus Ponens$)$
	
	\smallskip
	\noindent
	$-$ $\ALL = \{ \mathsf{A}, \mathsf{B}\}$ the set of algebras of argumentation labels where:\\
	
	\noindent $\mathsf{A}$ is an algebra of argumentation labels representing the trust degree attached to arguments. The domain of labels $A$ is the real interval $[0,1]$ representing a normalized trust valuation, where $\IES=1$ is the maximum valuation (and neutral element for $\OR$), while $\IE=0$ is the minimum valuation (and neutral element for \OA\ and \Con). Let $\alpha, \beta \in A$ be two labels, the operators of support, conflict and aggregation over labels representing the trust valuations associated with arguments are specified as follows:
	\begin{small}
		\begin{longtable}[c]{>{\arraybackslash}m{5.2cm}   >{\arraybackslash}m{5.8cm} }
			
			\footnotesize
			$\alpha \ \OR \ \beta  = \alpha\beta$
			&
			\small
			The support operator models the trust of a conclusion based on the conjunction of the trust valuations corresponding to the premises that support it.\\
			\\
			\
			
			\footnotesize
			$\alpha \ \OA \ \beta  = \alpha + \beta - \alpha \beta$
			&
			\small
			The aggregation operator states that if there is more than one argument for a conclusion, its trust valuation is the sum of the valuations of the arguments supporting it, with a penalty term.\\
			\\
			\
			
			\footnotesize
			$\alpha \ \OC \ \beta =$ $\left\{
			\begin{array}{lll}
			1& \mbox{ if } \alpha =1, \beta< 1\\[6pt]
			
			\displaystyle\frac{\alpha - \beta}{1 - \beta} & \mbox{ if } \alpha \geq \beta, \beta\neq 1\\[6pt]
			
			0& \mbox{ otherwise.}
			\end{array}
			\right. $
			&
			\small
			This conflict operator reflects that the trust valuation of a conclusion is weakened by the trust in its contrary.\\
		\end{longtable}
	\end{small}
	
	\noindent $\mathsf{B}$ is an algebra of argumentation labels representing a preference level attached to arguments. The domain of labels $B$ is again the real interval $[0,1]$ and represents a normalized preference valuation. The operations for $\mathsf{B}$ are specified as follows:
	\begin{small}
		\begin{longtable}[c]{>{\arraybackslash}m{5.2cm}    >{\arraybackslash}m{5.7 cm} }
			
			\footnotesize
			$\alpha \ \OR \ \beta  =  min(\alpha,\beta)$
			&
			\small
			The support operator reflects that an argument is as preferred as its weakest support, based on the weakest link rule.\\
			\\
			\
			\footnotesize
			$\alpha \ \OA \ \beta  = min(\alpha + \beta, 1)$
			&
			\small
			The aggregation operation reflects the idea that if we have more than one argument for a conclusion, its preference valuation is the sum of the preference valuations of the arguments that support it.\\
			\\
			\
			\footnotesize
			$\alpha  \OC  \beta = max(\alpha - \beta, 0)$
			&
			\small
			This conflict operation reflects that the valuation of a conclusion is weakened by the preference valuation of its contrary.\\
		\end{longtable}
	\end{small}
	
	\smallskip
	\noindent		
	$-$ \K\ is the following knowledge base; next to each presumption we show the trust and preference valuation associated by \F. These valuations are indicated between braces, where the first element represents the trust valuation and the second the preference valuation---valuations attached to rules represent the trust and preference of the connection between the antecedent and consequent of the rule. We annotate the attribute of an element with an interval $[x,y]$ to represent the variation of its attribute produced by the agent perception uncertainty. Rules are ground. However, following the usual convention\emph{~\cite{lifschitz1996foundations}}, some examples will use ``schematic rules'' with variables; to distinguish variables from other elements of a schematic rule, we will denote variables with an initial uppercase letter. We display below the set of formulas of \LLa\ forming \K:\\
	
	\begin{footnotesize}
		\begin{center}
			$\left\{
			\begin{array}{l}
			\tt{r}_1:\argdrule{\tt{buy}(X)}{\tt{goodArea}(X)} : 														\{0.85; 1\}\\[2pt]
			
			\tt{r}_2:\argdrule{\tt{buy}(X)}{\tt{goodOrientation}(X)} : 													\{1; 0.5\}\\[2pt]
			
			\tt{r}_3:\argdrule{\tt{goodArea}(X)}{\tt{basicServices}(X)} :												\{0.75; 0.95\}\\[2pt]
			\tt{r}_4:\argdrule{\tt{goodArea}(X)}{\tt{\tt{quietArea}(X)}} :												\{0.75; 0.9\}\\[2pt]
			\tt{r}_5:\argdrule{\tt{quietArea}(X)}{\tt{goodNeighbors}(X)}:												\{0.75; 0.9\}\\[2pt]
			\tt{r}_6:\argdrule{\tt{goodOrientation}(X)}{\tt{goodLight}(X),\tt{goodVentilation}(X)} : 					\{1; 0.5\}\\[2pt]
			\tt{r}_7:\argdrule{\tt{\nega{goodArea}(X)}}{\tt{insecureArea}(X)} : 										\{[0,0.5]; 1\}\\[2pt]
			\tt{r}_8:\argdrule{\tt{insecureArea}(X)}{\tt{gangOperate}(X)} : 											\{[0,0.5]; 1\}\\[2pt]	
			\tt{r}_9:\argdrule{\tt{\nega{insecureArea}}(X)}{\tt{reinforcePolice}(X)} : 									\{0.25; 1\}\\[2pt]		
			\tt{r}_{10}:\argdrule{\tt{\nega{gangOperate}}(X)}{\tt{\nega{insecureArea}}(X)} : 							\{0.25; 1\}\\[2pt]	
			\tt{r}_{11}:\argdrule{\tt{\nega{buy}(X)}}{\tt{highCostRenovation}(X)} : 									\{0.5; 0.65\}\\[2pt]					
			\tt{r}_{12}:\argdrule{\tt{highCostRenovation}(X)}{\tt{electricalProblem}(X), \tt{roofProblem}(X)} : 		\{1; 0.65\}\\[2pt]							
			\tt{r}_{13}:\argdrule{\tt{buy}(X)}{ \tt{goodArea}(X)} : 													\{0.5; 0.85\}\\[2pt]	
			\tt{r}_{14}:\argdrule{\tt{buy}(X)}{ \tt{goodConstruction}(X)} : 											\{1; 0.8\}\\[2pt]	
			\tt{r}_{15}:\argdrule{\tt{goodArea}(X)}{\tt{SafeArea}(X)} : 												\{0.5; 1\}\\[2pt]
			\tt{r}_{16}:\argdrule{\tt{safeArea}(X)}{\tt{reinforcePolice}(X)} :											\{[0,0.5]; 1\}\\[2pt]
			\tt{r}_{17}:\argdrule{\tt{goodArea}(X)}{\tt{lowPollution}(X)} : 											\{0.5; 0.85\}\\[2pt]				
			\tt{r}_{18}:\argdrule{\tt{goodConstruction}(X)}{\tt{qualityMaterials}(X)} : 									\{1; 0.6\}\\[2pt]	
			\tt{r}_{19}:\argdrule{\tt{goodConstruction}(X)}{\tt{adequateFooting}(X),\tt{solidFoundation}(X)} : 			\{1; 0.8\}\\[2pt]				
			\tt{r}_{20}:\argdrule{\nega{\tt{goodArea}(X)}}{\tt{precariousServices}(X)} : 								\{0.75; 0.95\}\\[2pt]	
			\tt{r}_{21}:\argdrule{\nega{\tt{buy}(X)}}{\tt{propertyTaxDebt}(X)} :										\{0.75; 0.5\}\\[2pt]			
			\end{array}
			\right\}$
			
			$\left\{
			\begin{array}{ll}
			\facto{\tt{basicService}(houseA)} : \{0.75; 0.95\}		& \facto{ \tt{roofProblem}(houseA)} : \{0.5; 0.6\} \\[2pt]	
			\facto{\tt{goodNeighbors}(houseA)} : \{0.75; 0.9\}	    & \facto{\tt{electricalProblem}(houseA)} :\{0.5; 0.5\} \\[2pt]	
			\facto{\tt{goodLight}(houseA)} : \{1; 0.4\} 			& \facto{\tt{reinforcePolice}(houseA)} : \{0.25; 1\}\\[2pt]
			\facto{\tt{goodVentilation}(houseA)} : \{1; 0.4\} 		& \facto{\tt{gangOperate}(houseA)} : \{[0,0.5]; 1\}\\[2pt]	
			\facto{ \tt{precariousServices}(houseB)} : \{0.75; 0.95\}  &  \facto{ \tt{lowAssaultLevel}(houseB)} : \{0.5; 0.1\} \\[2pt]			
			\facto{\tt{reinforcePolice}(houseB)} : \{[0,0.5]; 1\}  	   &  \facto{\tt{safe}(houseB)} : \{0.5; 1\}\\[2pt]
			\facto{\tt{lowPollution}(houseB)} : \{0.5; 0.85\}  		   &  \facto{ \tt{adequateFooting}(houseB)} : \{1; 0.7\} \\[2pt]
			\facto{\tt{solidFoundation}(houseB)} : \{1; 0.7\}  		   &  \facto{ \tt{propertyTaxDebt}(houseB)} : \{0.75; 0.5\} \\[2pt]
			\facto{\tt{qualityMaterials}(houseB)} : \{1; 0.6\}		   &
			\end{array}
			\right\}$
			
		\end{center}
	\end{footnotesize}
\end{Example}

We next present the notion of argumentation graph, which will be used to represent the argumentative analysis derived from an \textit{LAF}. We assume that there are no two nodes in a given graph that are named with the same sentence of \LLa, so we will use the naming sentence to refer to the I-node in the graph.
\begin{Definition}[Argumentation Graph]~\label{Def.ArgumentationGraph}
	Given an LAF $\Phi = $ \LAFARG\, its associated  argumentation graph is the digraph $\Graph{\Phi} = (N, E)$, where $N \neq \emptyset$ is the set of nodes and $E$ is the set of the edges, constructed as follows:
	
	\begin{itemize}
		\item[i\emph{)}] each element $X \in \K$ or derived from \K\ through \R, is represented by an I-node $X \in N$; there must not exist two distinct I-nodes $X,Y\in N$ representing the same formula $\varphi$.
		\item[ii\emph{)}] for each application of an inference rule defined in $\Phi$, there exists an RA-node $R \in N$ such that:
		\begin{enumerate}
			\item[--] the inputs are all I-nodes $P_1,\ldots, P_m \in N$ representing the premises necessary for the application of the rule $R$ (including the defeasible rules of \K\ that may be applied).
			\item[--] the output is an I-node $Q \in N$ representing the conclusion.
		\end{enumerate}
		\item[iii\emph{)}] if $X$ and $\negaBar{X}$ are in $N$, then there exists a CA-node with edges to and from both I-nodes $X$ and $\negaBar{X}$.	
		
		\item[iv\emph{)}] for all $X \in N$ there does not exist a path from $X$ to $X$ in $G_{\Phi}$ that does not pass through a CA-node. That is, $G_{\Phi}$ without the CA-nodes is acyclic.
	\end{itemize}
	
	Condition~(iv) forbids cycles other than the mutual attacks between nodes. This seems to be too restrictive, but it must be noted that RA-node cycles are mostly generated by faulty specifications.
\end{Definition}

Given an argumentation graph, it is possible to identify different subgraphs representing the arguments that participate in the dialectical process. Briefly speaking, an argument for a formula $\varphi$ (represented as an I-node in the graph $G_{\Phi}$) is a subgraph of $G_{\Phi}$ representing the different reasons supporting such formula. Formally:

\begin{Definition}[Argument]\label{def.argumentoLAF}
	Let $\Phi= \LAFARG$ be an LAF, and $G_\Phi$ be the associated argumentation graph without CA-nodes for $\Phi$. An argument for a formula represented through an I-node $X$ is the subgraph $\STR{A} = (N',E')$ of $G_\Phi$ composed by all the ancestors of $X$, such that $N' \neq \emptyset$ and $E' \neq \emptyset$. We will denote with $\textit{Arg}_\Phi$ the set of all arguments that can be identified in $G_\Phi$.
\end{Definition}

An argumentation graph without CA-nodes can be interpreted like an \emph{acyclic directed graph} (DAG), since there is a unique I-node representing a formula of $\K$ or derived from \K\ through \R, and the correct foundation of every formula represented through an I-node in $G_{\Phi}$ (there exist no cycles produced by the application of inference rules supporting a formula in $G_{\Phi}$). In consequence, by Definition~\ref{def.argumentoLAF}, an argument that belongs to an argumentative graph $G_\Phi$ is represented by an acyclic directed graph. This representation allows us to naturally model accrued arguments~\cite{verheijaccrual,AcrualPrakken,Lucero13}, since such arguments are composed by all the support chains that exist for their corresponding conclusion. For that, in \emph{LAF}, the strength of an argument is represented by the aggregation of the strengths associated with each independent reasoning chain supporting the conclusion. Also, the aggregation of arguments reduces the complexity of conflict resolution, since there exists a single point of conflict in $G_{\Phi}$ for each pair of contradictory formulas. Therefore, it will be possible to implement a weakening relation between contradictory arguments, since the weakened strength does not depend on the order in which the conflicts are taken into account.

Next we will present an example of an argumentative graph $G_{\Phi}$, identifying the different subgraphs that represent the arguments.

\begin{Example}
	\label{Example4}
	Applying the inference rules defined in \R\ over the knowledge base \K\ presented in Example~\ref{Example_InstantiateLAF}, we get the argumentation graph $\Graph{\Phi}$ presented in Figure~\ref{Fig.argumentationgraph}, which is organized in two groups; the first gives reasons for and against buying house~A, while the second group shows reasons for and against buying house~B.
	
	For the first group (in the center) there are two inference chains supporting the claim $\tt{buy}(houseA)$---one contains information about house~A's area taking into account services and residents, and the other talks about house~A's orientation considering natural light and ventilation. On the right, there is one inference chain supporting the claim $\nega{\tt{buy}(houseA)}$ based on construction problems. On the left, we present an inference chain justifying the claim $\nega{\tt{goodArea}(houseA)}$, where the agent considers house~A's insecurity; however, house~A's security is reinforced when taking into account police activity, giving us reasons to consider house~A's area as secure.
	For the second group (in the center) there are two inference chains supporting the claim $\tt{buy}(houseB)$---one contains information about house~B's area taking into account security and pollution levels, and the second references reasons regarding the good construction of house~B. On the left, we present an inference chain justifying the claim $\nega{\tt{goodArea}(houseB)}$ based on the precarious services available in house~B's area. On the right, there is one inference chain supporting the claim $\nega{\tt{buy}(houseB)}$ based on property tax debt. Note that CA-nodes link complementary literals.
\end{Example}

\afterpage{
	\begin{landscape}
		\begin{figure}[p]
			\centering
			\includegraphics[width=1.8\textwidth]{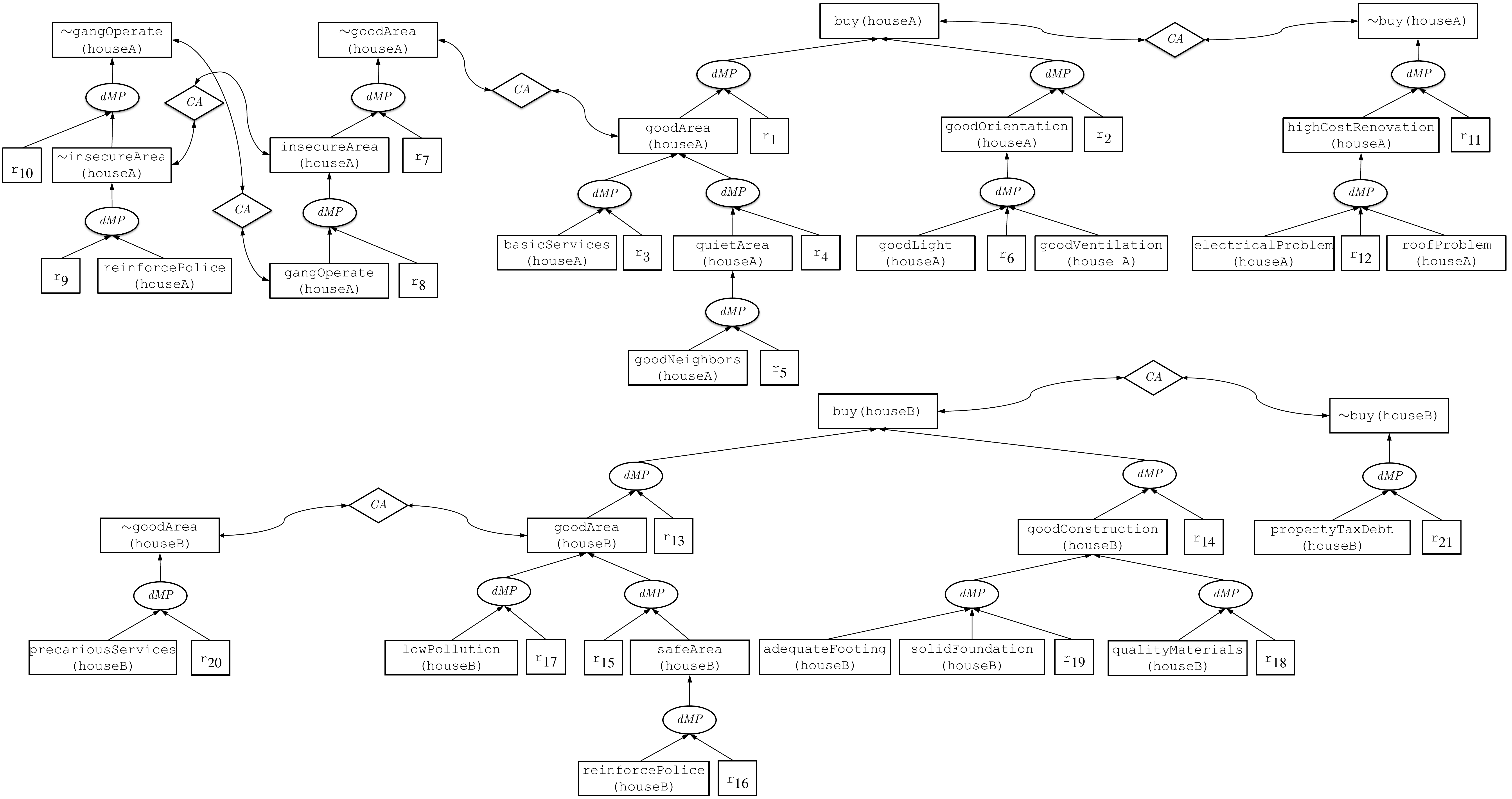}
			\caption{Argumentation graph for Example~\ref{Example4}}\label{Fig.argumentationgraph}
		\end{figure}
	\end{landscape}
}

\subsection{Labeling Procedure for a Labeled Argumentation Framework}

Once the argumentation graph is obtained, we proceed to attach a label to each I-node, representing valuations referring to extra information that we want to represent. Each feature of an I-node is represented through an algebra $\mathsf{A}_i$ of $\mathcal{A}$, assigning two valuations \Labelplus{i}{X} and \Labelminus{i}{X}, where \Labelplus{i}{X} represents the \emph{accrued valuation}, obtained through the accrual and support operations defined in algebra $\mathsf{A}_i$, while \Labelminus{i}{X} is the \emph{weakened valuation}, obtained through the conflict operation. Next, we present the labeling procedure for an argumentation graph, which derives a {\em system of equations} that characterizes the knowledge contained in the formalism.

\begin{Definition}[Labeling Procedure for a Graph]~\label{Def.Labelgraphcycle}
	Let $\Phi= \LAFARG$ be an LAF, and $\Graph{\Phi}$ be its corresponding argumentation graph.
	Let $\mathsf{A}_i$ be one of the algebras in $\mathcal{A}$,  representing a feature to be associated with each I-node $X$.
	A labeled argumentation graph is an assignment of two valuations from each of the algebras to all I-nodes of the graph, denoted with \Labelplus{i}{X} and \Labelminus{i}{X}, where \Labelplus{i}{X} accounts for the aggregation of the reasons supporting the claim $X$, while \Labelminus{i}{X} displays the state of the claim after taking conflict into account, such that \Labelplus{i}{X}, \Labelminus{i}{X} $\in \mathit{A}_i$. If $X$ is an I-node, its valuations are determined as follows:
	
	\begin{itemize}
		\item[i\textup{)}]  If $X$ has no inputs, then its accrued valuation is given by function \F; thus, we define  $\Labelplus{i}{X} = \F_i(X)$.
		
		\item[ii\textup{)}] If $X$ has an input from a CA-node representing conflict with an I-node $\negaBar{X}$, then:
		$\Labelminus{i}{X} =\Labelplus{i}{X} \Con\ \Labelplus{i}{\overline{X}}$.
		
		If there is no input from a CA-node, then: $\Labelminus{i}{X} = \Labelplus{i}{X}$.
		
		\item[iii\textup{)}] If $X$ is an element of $\K$ with inputs from RA-nodes $R_1, \ldots, R_k$, where each $R_s$ has premises $X_{1}^{R_s}, \ldots, X_{n_s}^{R_s}$, then:
		$$\Labelplus{i}{X} = \F_i(X) \Agr \left[\Agr^k_{s=1}\left(\Sup^{n_s}_{t=1} \Labelminus{i}{X_t^{R_s}}\right)\right]$$
		
		If $X$ is not an element of $\K$ and has inputs from RA-nodes $R_1, \ldots, R_k$, where each $R_s$ has premises $X_{1}^{R_s}, \ldots, X_{n_s}^{R_s}$, then:
		
		$$\Labelplus{i}{X} = \Agr^k_{s=1}\left(\Sup^{n_s}_{t=1} \Labelminus{i}{X_t^{R_s}}\right)$$
		
		%		To get this equation, we first use the support operation applied to the weakened valuations assigned to the premises of each % of the rules $R_s$ that form the body of an argument supporting $X$, and then we calculate the accrual of all these arguments.
	\end{itemize}
	
	\noindent
	The label of an I-node $X$ is then an n-tuple of pairs of valuations
	$\left((\Labelplus{1}{X},\Labelminus{1}{X}),\right.$ \linebreak $\left.(\Labelplus{2}{X}, \Labelminus{2}{X}), \ldots, (\Labelplus{i}{X},\Labelminus{i}{X}), \ldots, (\Labelplus{n}{X}, \Labelminus{n}{X})\right)$, where each pair $(\Labelplus{i}{X},\Labelminus{i}{X})$ represents the valuations of a certain feature associated with algebra $\algebra{A}_i \in \mathcal{A}$.
\end{Definition}

\begin{figure*}[ht!]
	\begin{center}
		\includegraphics[width=.9\textwidth]{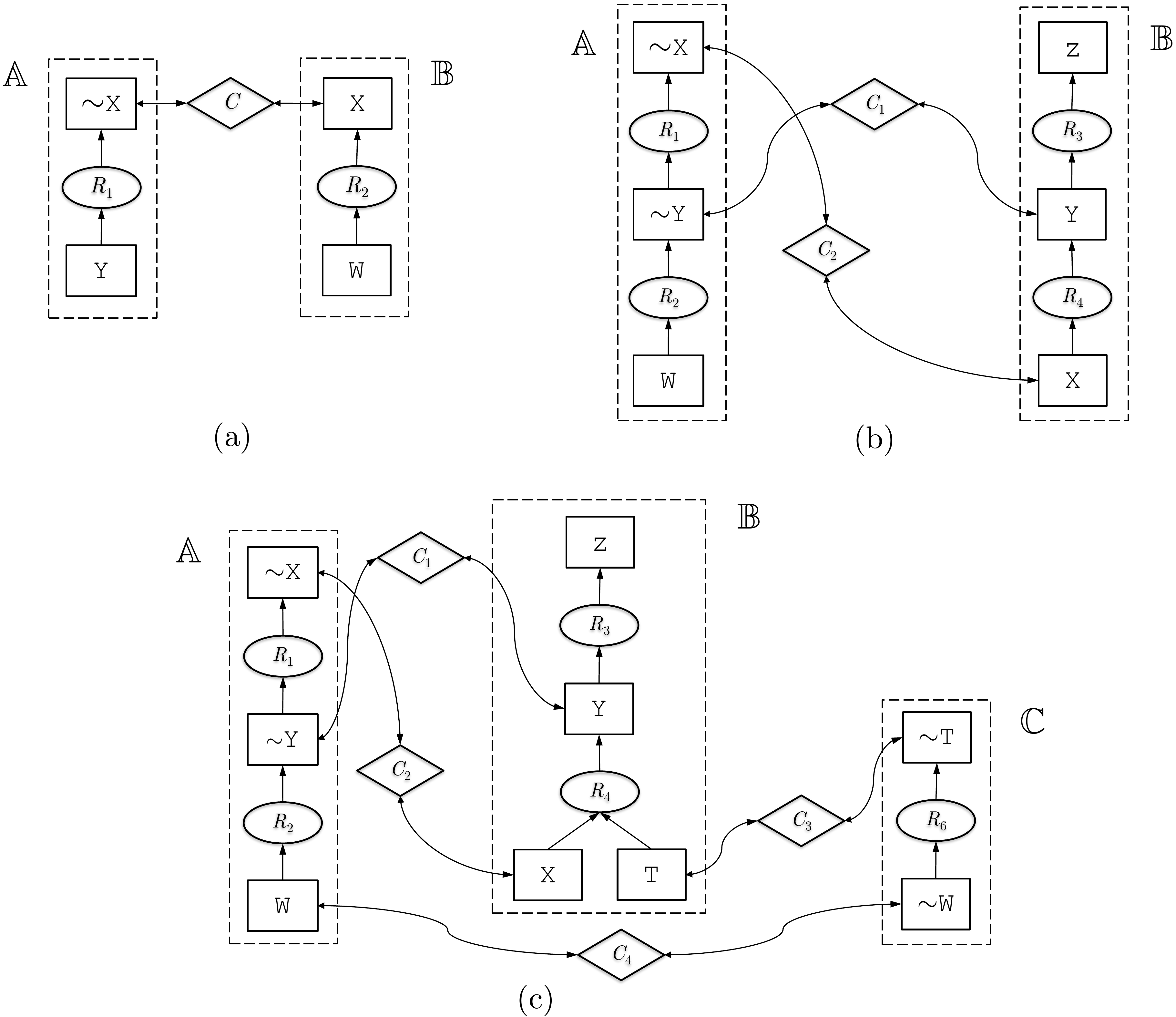}
		\caption{Cycles in an argumentation graph involving CA-nodes}\label{Fig.cicloCAnodos}
	\end{center}
\end{figure*}

In LAF, an argumentative graph can contain cycles (cf.\ Figure~\ref{Fig.cicloCAnodos}) produced by one or more conflicts between two or more arguments. These conflicts model the inconsistencies present in the knowledge about the real world. Therefore, the labeling procedure must determine the restrictions that the valuations must fulfill by adequately considering these inconsistencies.

There are different kinds of conflict cycles---the simplest is the one shown in Figure~\ref{Fig.cicloCAnodos}-(a), where we identify a conflict relation between two arguments that support contradictory conclusions. This kind of cycle does not have a large impact on the computational cost, since it is possible to obtain the accrued valuations for each argument independently of each other and finally obtain the weakened valuations associated with all the involved conclusions; in this sense, the conflict resolution for this pair of arguments has a minimal dependence. However, more complex conflict cycles can arise, increasing the cost of finding their solution. Figure~\ref{Fig.cicloCAnodos}-(b) illustrates another typical case in which two arguments, \STR{A} and \STR{B}, have a blocking conflict---i.e., the resolution of the conflicts involved are dependent on each other: the weakened valuation associated with the premise $\mathtt{X}$ of \STR{B} depends on the accrued valuation assigned to the conclusion $\nega{\mathtt{X}}$ of \STR{A}, which is supported by the weakened valuation of an intermediate premise $\nega{\mathtt{Y}}$. To obtain the weakened valuation of $\nega{\mathtt{Y}}$, it is necessary to resolve the conflict with the intermediate premise $\mathtt{Y}$ of \STR{B}, which has an accrued valuation obtained through the weakened valuation of the premise claim $\mathtt{X}$ that supports it.
Finally, Figure~\ref{Fig.cicloCAnodos}-(c) shows a case in which conflict cycles are combined, and the resolution of conflicts between arguments \STR{A} and \STR{B} depend on the conflict resolution between the arguments \STR{B} and \STR{C}, and \STR{A} and \STR{C}---the weakened valuation associated with the premise $\mathtt{X}$ of \STR{B} depends on the accrued valuation assigned to the conclusion $\nega{\mathtt{X}}$ of \STR{A}, which is supported by the weakened valuation of an intermediate premise $\nega{\mathtt{Y}}$. First, to obtain the weakened valuation of $\nega{\mathtt{Y}}$ it is necessary to calculate its accrued valuation through the weakened valuation associated with the premise that supports such formula (the claim $\nega{\mathtt{W}}$) and establish the accrued valuation of the contrary claim $\mathtt{Y}$ specifying the weakened valuation for the premises $\mathtt{X}$ and $\mathtt{T}$ that support such claim. Once the weakened valuation associated with $\nega{\mathtt{W}}$ of \STR{C} is obtained, we determine the accrued valuation of the conclusion for such argument (claim $\nega{\mathtt{T}}$). Only then can the conflict between the claims $\mathtt{T}$ and $\nega{\mathtt{T}}$ be resolved by obtaining (through function \F) the accrued valuation of $\mathtt{T}$, premise of argument \STR{A}. The weakened valuation $\nega{\mathtt{W}}$ is obtained by resolving the conflict with the claim $\mathtt{W}$, where both claims have an accrued valuation obtained through function \F. In short, these cases show that the computational cost associated with the process of resolving conflicts between arguments is related to the dependencies that they establish to calculate their accrued and weakened valuations.

Our labeling procedure determines the system of equations representing the constraints that all valuations must fulfill, i.e., each solution is a possible configuration of values that stabilizes the propagation of the attributes of the knowledge in the graph. Intuitively, this problem is akin to a problem in linear or non linear programming, depending on the operations chosen and the number of variables determined by the number of conflict cycles involved---if the knowledge behavior is captured through a linear system, then linear programming techniques can be applied to analyze and solve this system~\cite{khachiyan1980polynomial,karmarkar1984new}; on the other hand, if the system is not linear, then non-linear programming techniques (with higher complexity) can be applied~\cite{harker1990finite,simon2010kalman}. Linear systems are generally obtained when certain mathematical assumptions or approximations are made that allow for easier calculations; non-linear systems, however, are usually hard (or impossible) to treat, and their behavior with respect to a given variable is not easy to predict. Furthermore, some non-linear systems have an integrated or exact solution, while others are impossible to solve. In future work we will advance in the characterization of the algebra of argumentation labels, seeking to establish conditions that ensure the creation of a system of equations that is both solvable and computationally tractable.

Next, we present the algorithms for obtaining the valuations associated with each I-node of an argumentation graph. Algorithm~1  produces a solution for a system of equations by using a solver---the specific solver used will depend on the operators defined in the algebra. Algorithm~2 analyzes each node of the graph with the purpose of specifying the equations that determine the valuations of its I-nodes. To do this, it first analyzes the chains of inference rules that support an I-node to determine the support/aggregation valuation associated with it; then, it specifies the equation that determines the weakened valuation of the I-node $X$ based on the support/aggregation valuation of $\negaBar{X}$. Note that the propagation of the characteristics through a reasoning chain is based on the weakened valuation associated with each I-node that comprises such chain, giving rise to a dependence in the argumentation graph propagation.\\

\IncMargin{0em}
\begin{algorithm}[ht]
	\small	
	\label{algorithm_1}
	\caption{Labeling procedure for a graph $\Graph{\Phi}$}
	\KwIn{Argumentation graph $\Graph{\Phi}$ and knowledge base \K.}
	\KwOut{System of equations $\textit{EQS}$ for valuations \Labelplus{i}{X} and \Labelminus{i}{X} associated with the I-nodes $X_i$ of $\Graph{\Phi}$.}
	Initialize $\textit{EQS}$ as an empty system of equations\;
	Initialize each I-node of $\Graph{\Phi}$ with a \emph{not visited status}\;
	\For{each I-node $X$ in $G$ that is not visited}
	{	
		$\textit{EQS} := \textit{EQS} \cup \textit{LabelingFunction}(X, \Graph{\Phi},\K, \textit{EQS})$\;
	}
	\Return $\textit{EQS}$.
\end{algorithm}	
\DecMargin{0em}

\IncMargin{0em}
\begin{algorithm}[ht]
	\small
	\caption{Labeling Function for nodes in the graph}
	\label{algorithm_2}
	\KwIn{Argumentation graph $\Graph{\Phi}$, I-node $X$, and knowledge base \K.}
	\textbf{Input/Output}: System of equations $\textit{EQS}$ for valuations \Labelplus{i}{X} and \Labelminus{i}{X} associated with the I-nodes $X_i$ of $\Graph{\Phi}$.
	
	\vskip4pt
	Mark the I-node $X$ as visited\;
	\eIf{X has input from RA-nodes}
	{
		\eIf{$X$ is an element of \K}
		{
			$\textit{Aggregation} := \F(X)$\;
		}
		{
			$\textit{Aggregation} := \bot_i$\;
		}
		\For{each RA-node $R$ with an incoming edge from $X$}
		{	
			$\textit{Support} := \top_i$\;
			\For{for each premise $P$ with outgoing edge to an RA-node $R$}
			{
				\eIf{P is not visited}
				{
					$\textit{LabelingFunction}(P, \Graph{\Phi},\K, \textit{EQS})$\;
				}
				{
					$\textit{Support} := \textit{Support} \Sup \Labelminus{i}{P}$\;
				}
			}
			$\textit{Aggregation} := \textit{Aggregation} \OA \textit{Support}$;	
		}									
		Add ``$\Labelplus{i}{X} = \textit{Aggregation}$'' to $\textit{EQS}$\;
	}
	{
		Add ``$\Labelplus{i}{X} = \F(X)$'' to $\textit{EQS}$\;
	}	
	\eIf{X has input from CA-nodes}
	{
		\eIf{\negaBar{X} is not visited}
		{
			$\textit{LabelingFunction}(\negaBar{X}, \Graph{\Phi}, \K)$\;
		}
		{
			Add ``$\Labelminus{i}{X} = \Labelplus{i}{X} \Con \Labelplus{i}{\negaBar{X}}$'' and
			``$\Labelminus{i}{\negaBar{X}} = \Labelplus{i}{\negaBar{X}} \Con \Labelplus{i}{X}$'' to $\textit{EQS}$\;
		}
	}		
	{		
		Add ``$\Labelminus{i}{X} = \Labelplus{i}{X}$'' to $\textit{EQS}$\;
	}
	\Return $\textit{EQS}$.
\end{algorithm}	
\DecMargin{0em}

The following result states the computational cost of this procedure.

\begin{Proposition}
	The worst-case running time of Algorithm~1 is $O(m\times t)$, where $m$ is the number of I-nodes and $t$ is the
	number of RA-nodes in the graph.
\end{Proposition}

The semantics of an argumentation graph is determined by the possible solutions to the system of equations generated by the labeling process---each solution is a \emph{valid labeling} for the argumentation graph, representing a \emph{model}. Formally, we have:

\begin{Definition}[Valid Labeling in an LAF]~\label{def.modeleqs}
	Let $\Phi= \langle \mathcal{L}, \mathcal {R},$  $\mathcal{K}, \mathcal{A}, \mathcal{F} \rangle$ be an LAF, $\Graph{\Phi}$ be the corresponding argumentation graph for $\Phi$, and $\textit{EQS}$ be the corresponding system of equations
	representing the constraints that all the valuations associated with the knowledge of $\Graph{\Phi}$ must fulfill.
	A {\em valid labeling} for $\Graph{\Phi}$, denoted \model, is any set of values $\Labelplus{i}{X_i}$ and $\Labelminus{i}{X_i}$ that constitute a solution to $\textit{EQS}$.
\end{Definition}

\begin{Theorem}[Convergence]\label{teom.convergencia}
	Let $\Phi= \LAFARG$ be an LAF, $G_\Phi$ be the corresponding argumentation graph for $\Phi$, and $S_1$ and $S_2$ be two different labeling sequences for $G_\Phi$, where $S_1$ generates system $EQS_1$ and $S_2$ generates system $EQS_2$.
	Then, we have that $EQS_1 = EQS_2$.
\end{Theorem}

As we mentioned before, the supported and weakened valuations of an I-node have a strong relationship. The valuations associated with each I-node in the graph corresponding to a valid labeling satisfy the following properties.

\begin{Lemma}\label{prop.values}
	Let $\Phi= \LAFARG$ be an LAF, $G_\Phi$ be the corresponding argumentation graph for $\Phi$, \model\ be a valid labeling for $G_{\Phi}$, and $X$ be an I-node in $G_{\Phi}$. Then, the labels \Labelplus{i}{X} and \Labelminus{i}{X} representing each feature of $X$ through an algebra $\mathsf{A}_i$ in $\mathcal{A}$ satisfy the following properties:
	\begin{itemize}
		\item[i)]  $\Labelminus{i}{X} \leq \Labelplus{i}{X}$ (thus, if $\Labelplus{i}{X}=\bot$, it follows that $\Labelminus{i}{X} =\bot$ as well).
		
		\item[ii)]  If $\Labelplus{i}{\overline{X}} = \IE$, then $\Labelminus{i}{X} = \Labelplus{i}{X}$.
		
		\item[iii)]  If $\Labelplus{i}{X} \leq \Labelplus{i}{\overline{X}}$, then $\Labelminus{i}{X} =\bot$.
	\end{itemize}
\end{Lemma}

In addition, the following underlying principles are satisfied by all the valuations defined according to a model for a particular system of equations. In general, these principles describe the behavior of valuations associated with arguments proposed by different works in literature, such as those by Cayrol and Lagasquie-Schiex~\cite{cayrol2005graduality}, Jakobovits and Vermeir~\cite{jakobovits1999robust}, Besnard and Hunter~\cite{besnard2001logic}, and Dunne \emph{et.al.}~\cite{dunne2011weighted}, among others.

\begin{Property}[Underlying Principles]\label{prop.principlesvaluations}
	The valuations given by Definition~\ref{Def.Labelgraphcycle} respect the following principles:
	\begin{itemize}
		\item[$P_1$] The weakened valuation is equal to the supported/accrued valuation for the arguments without attackers, and for an attacked but undefeated argument the weakened valuation is less than the supported/accrued valuation, if the attacking arguments are strong enough to weaken it.
		
		\item[$P_2$] The weakened valuation for an argument depends, in a non-increasing manner, on the supported/accrued valuation of the attacking argument.
		
		\item[$P_3$] The valuations of arguments supporting a claim $X$ contribute to increase the accrued valuation of $X$, and for this reason they increase the strength of the direct attack to the argument that supports the opposite claim $\overline{X}$.	
	\end{itemize}
\end{Property}

Once all possible models for $\textit{EQS}$ are determined, a user might be interested in analyzing this set of models with the purpose of optimizing the features associated with a specific set of elements, taking in this way a subspace of possible solutions where their own requirements are fulfilled.
In this sense, the problem is similar to a mathematical optimization problem consisting of an objective function and a set of constraints expressed in the form of a system of equations characterizing a specific situation. A feasible solution that minimizes (or maximizes) some measure is called an \emph{optimal solution}. Towards this end, it is possible to introduce objective functions $\textit{maximize}(\Labelplus{}{\tt{X}})$ or $\textit{minimize}(\Labelplus{}{\tt{X}})$ in our system of equations $\textit{EQS}$ obtaining a set of optimal solutions that represents the point of view of a particular user. Clearly, since we are simply adding an objective function, the set of solutions of the new system is a subset of the solutions for a system with no such function.

Optimization models are used extensively in almost all areas of decision-making, such as engineering design, medical procedures, legal cases analysis, and financial portfolio selection. In our example, the trust degree associated with the source ``neighbors'', which offers arguments related to the security of the houses' area, can oscillate between $0$ and $0.5$. Then, if the agent is interested in optimizing the trust degree for the source ``neighbors'', a subspace of the possible solutions where this trust degree is maximized (or minimized) is obtained. Next, we determine the system of equations for the argumentation graph $G$ associated with our instantiation of $\Phi$ (Example~\ref{Example4}) according to the labeling procedure. In addition, we extend the system of equations in order to maximize and minimize the trust degree of the knowledge pieces offered by the sources \emph{``neighbors''}.

\begin{Example}\label{Ex.LabelArgumentationGraph}
	Consider again the setup from Example~\ref{Example_InstantiateLAF} and the argumentation graph presented in Example~\ref{Example4}. Figure~\ref{fig:EQS} shows the system of equations $\textit{EQS}$ that represents the constraints for this LAF instance.
	Note that, for reasons of readability, we do not include the equations that determine the valuations of the leaf nodes, which are trivial.
	
	In Figure~\ref{Fig.labeledargumentationgraph} we depict the solutions to $\textit{EQS}$, which represent the models of the instantiated LAF. In this case, we specify the possible models for $\Phi$ maximizing the trust degree of the formulas offered by the source \emph{``house A's neighbors''} and minimizing the trust degree of those offered by the source \emph{``house B's neighbors''}.
\end{Example}	

\begin{figure}[p]
	\begin{center}
		$\left\{
		\begin{array}{l}
		\tt{e}_1: \Labelplus{}{\tt{\nega{insecureArea}}(houseA)} = \Labelminus{}{\tt{reinforcePolice}(houseA)} \ \Sup \ \Labelminus{}{\tt{r}_{9}}\\[2pt]
		\tt{e}_2: \Labelminus{}{\tt{\nega{insecureArea}}(houseA)} = \Labelplus{}{\tt{\nega{insecureArea}}(houseA)} \ \Con \ \Labelplus{}{\tt{insecureArea}(houseA)} \\[2pt]
		\tt{e}_3: \Labelplus{}{\tt{\nega{gangOperate}}(houseA)} = \Labelminus{}{\tt{\nega{insecureArea}}(houseA)} \ \Sup \ \Labelminus{1}{\tt{r}_{10}}\\[2pt]
		\tt{e}_4: \Labelminus{}{\tt{\nega{gangOperate}}(houseA)} = \Labelplus{}{\tt{\nega{gangOperate}}(houseA)} \ \Con \ \Labelplus{}{\tt{gangOperate}(houseA)} \\[2pt]
		\tt{e}_5: \Labelminus{}{\tt{gangOperate}(houseA)} = \Labelplus{}{\tt{gangOperate}(houseA)} \ \Con \ \Labelplus{}{\tt{\nega{gangOperate}}(houseA)} \\[2pt]
		\tt{e}_6: \Labelplus{}{\tt{insecureArea}(houseA)} = \Labelminus{}{\tt{gangOperate}(houseA)} \ \Sup \ \Labelminus{}{\tt{r}_{8}}\\[2pt]
		\tt{e}_7: \Labelminus{}{\tt{insecureArea}(houseA)} = \Labelplus{}{\tt{insecureArea}(houseA)} \ \Con \ \Labelplus{}{\tt{\nega{insecureArea}}(houseA)} \\[2pt]
		\tt{e}_8: \Labelplus{}{\tt{\nega{goodArea}}(houseA)} = \Labelminus{}{\tt{insecureArea}(houseA)} \ \Sup \ \Labelminus{}{\tt{r}_{7}}\\[2pt]
		\tt{e}_9: \Labelminus{}{\tt{\nega{goodArea}}(houseA)} = \Labelplus{}{\tt{\nega{goodArea}}(houseA)} \ \Con \ \Labelplus{}{\tt{goodArea}(houseA)} \\[2pt]
		\tt{e}_{10}: \Labelplus{}{\tt{goodArea}(houseA)} = \Labelminus{}{\tt{basicServices}(houseA)} \ \Sup \ \Labelminus{}{\tt{r}_{4}}\\[2pt]
		\tt{e}_{11}: \Labelminus{}{\tt{goodArea}(houseA)} = \Labelplus{}{\tt{goodArea}(houseA)} \ \Con \ \Labelplus{}{\tt{\nega{goodArea}}(houseA)} \\[2pt]
		\tt{e}_{12}: \Labelplus{}{\tt{quietArea}(houseA)} = \Labelminus{}{\tt{goodNeighbors}(houseA)} \ \Sup \  \Labelminus{}{\tt{r}_{5}}\\[2pt]
		\tt{e}_{13}: \Labelminus{}{\tt{quietArea}(houseA)} = \Labelplus{}{\tt{quietArea}(houseA)}\\[2pt]
		\tt{e}_{14}: \Labelplus{}{\tt{goodArea}(houseA)} = \Labelminus{}{\tt{goodOrientation}(houseA)} \ \Sup \ \Labelminus{}{\tt{r}_{2}}\\[2pt]
		\tt{e}_{15}: \Labelplus{}{\tt{goodOrientation}(houseA)} = \Labelminus{}{\tt{goodLight}(houseA)} \ \Sup \ \Labelminus{}{\tt{goodVentilation}(houseA)} \ \Sup \ \Labelminus{}{\tt{r}_{6}}\\[2pt]
		\tt{e}_{16}: \Labelminus{}{\tt{goodOrientation}(houseA)} = \Labelplus{}{\tt{goodOrientation}(houseA)}\\[2pt]
		\tt{e}_{17}: \Labelplus{}{\tt{buy}(houseA)} = (\Labelminus{}{\tt{goodArea}(houseA)} \ \Sup \ \Labelminus{}{\tt{r}_{1}}) \ \Agr \ (\Labelminus{}{\tt{goodOrientation}(houseA)} \ \Sup \ \Labelminus{}{\tt{r}_{2}})\\[2pt]
		\tt{e}_{18}: \Labelminus{}{\tt{buy}(houseA)} = \Labelplus{}{\tt{buy}(houseA)} \ \Con \ \Labelplus{}{\tt{\nega{buy}}(houseA)} \\[2pt]
		\tt{e}_{19}: \Labelminus{}{\tt{\nega{buy}}(houseA)} = \Labelplus{}{\tt{\nega{buy}}(houseA)} \ \Con \ \Labelplus{}{\tt{buy}(houseA)} \\[2pt]
		\tt{e}_{20}: \Labelplus{}{\tt{\nega{buy}}(houseA)} = \Labelminus{}{\tt{highCostRenovation}(houseB)} \ \Sup \ \Labelminus{}{\tt{r}_{11}}\\[2pt]
		\tt{e}_{21}: \Labelminus{}{\tt{highCostRenovation}(houseA)} = \Labelplus{}{\tt{highCostRenovation}(houseA)}\\[2pt]
		\tt{e}_{22}: \Labelplus{}{\tt{highCostRenovation}(houseA)} = \Labelminus{}{\tt{electricalProblem}(houseA)} \ \Sup \  \Labelminus{}{\tt{roofProblem}(houseA)} \ \Sup \ \Labelminus{}{\tt{r}_{12}}\\[2pt]
		\tt{e}_{23}: \Labelplus{}{\tt{\nega{goodArea}}(houseB)} = \Labelminus{}{\tt{precariousServices}(houseB)} \ \Sup \ \Labelminus{}{\tt{r}_{20}}\\[2pt]
		\tt{e}_{24}: \Labelminus{}{\tt{\nega{goodArea}}(houseB)} = \Labelplus{}{\tt{\nega{goodArea}}(houseB)} \ \Con \ \Labelplus{}{\tt{goodArea}(houseB)} \\[2pt]
		\tt{e}_{25}: \Labelplus{}{\tt{goodArea}(houseB)} = (\Labelminus{}{\tt{lowPollution}(houseB)} \ \Sup \ \Labelminus{}{\tt{r}_{17}})\ \Agr \
		(\Labelminus{}{\tt{safeArea}(houseB)} \ \Sup \ \Labelminus{}{\tt{r}_{15}})\\[2pt]
		\tt{e}_{26}: \Labelminus{}{\tt{safeArea}(houseB)} = \Labelplus{}{\tt{safeArea}(houseB)}\\[2pt]
		\tt{e}_{27}: \Labelplus{}{\tt{safeArea}(houseB)} =  (\Labelminus{}{\tt{reinforcePolice}(houseB)} \ \Sup \ \Labelminus{}{\tt{r}_{16}}) \ \Agr \ \F(\footnotesize{\text{\tt{safeArea}(houseB)}})\\[2pt]
		\tt{e}_{28}: \Labelminus{}{\tt{goodConstruction}(houseB)} = \Labelplus{}{\tt{goodConstruction}(houseB)}\\[2pt]
		\tt{e}_{29}: \Labelplus{}{\tt{goodConstruction}(houseB)} = (\Labelminus{}{\tt{adequateFooting}(houseB)} \ \Sup \ \Labelminus{}{\tt{solidFoundation}(houseB)} \ \Sup \ \Labelminus{}{\tt{r}_{19}}) \ \Agr \ \\[2pt](\Labelminus{}{\tt{qualityMaterials}(houseB)} \ \Sup \ \Labelminus{}{\tt{r}_{18}})\\[2pt]
		\tt{e}_{30}: \Labelplus{}{\tt{buy}(houseB)} = (\Labelminus{}{\tt{goodArea}(houseB)} \ \Sup \ \Labelminus{}{\tt{r}_{13}}) \ \Agr \ (\Labelminus{}{\tt{goodConstruction}(houseB)} \ \Sup \ \Labelminus{}{\tt{r}_{14}})\\[2pt]
		\tt{e}_{31}: \Labelminus{}{\tt{buy}(houseB)} = \Labelplus{}{\tt{\tt{buy}(houseB)}} \ \Con \ \Labelplus{}{\nega\tt{buy}(houseB)} \\[2pt]
		\tt{e}_{32}: \Labelplus{}{\nega\tt{buy}(houseB)} = \Labelminus{}{\tt{propertyTaxDebt}(houseB)} \ \Sup \ \Labelminus{}{\tt{r}_{21}}\\[2pt]
		\tt{e}_{33}: \Labelminus{}{\nega\tt{buy}(houseB)} = \Labelplus{}{\nega\tt{\tt{buy}(houseB)}} \ \Con \ \Labelplus{}{\tt{buy}(houseB)} \\[2pt]
		\tt{e}_{34}: max(\Labelplus{}{\tt{gangOperate}(houseA)})\\[2pt]
		\tt{e}_{35}: min(\Labelplus{}{\tt{reinforcePolice}(houseB)})\\[2pt]
		\end{array}
		\right\}$
	\end{center}
	\caption{System of equations for Example~\ref{Ex.LabelArgumentationGraph}.}
	\label{fig:EQS}
\end{figure}
\afterpage{
	\begin{landscape}
		\begin{figure}[p]
			\centering
			\includegraphics[width=1.8\textwidth]{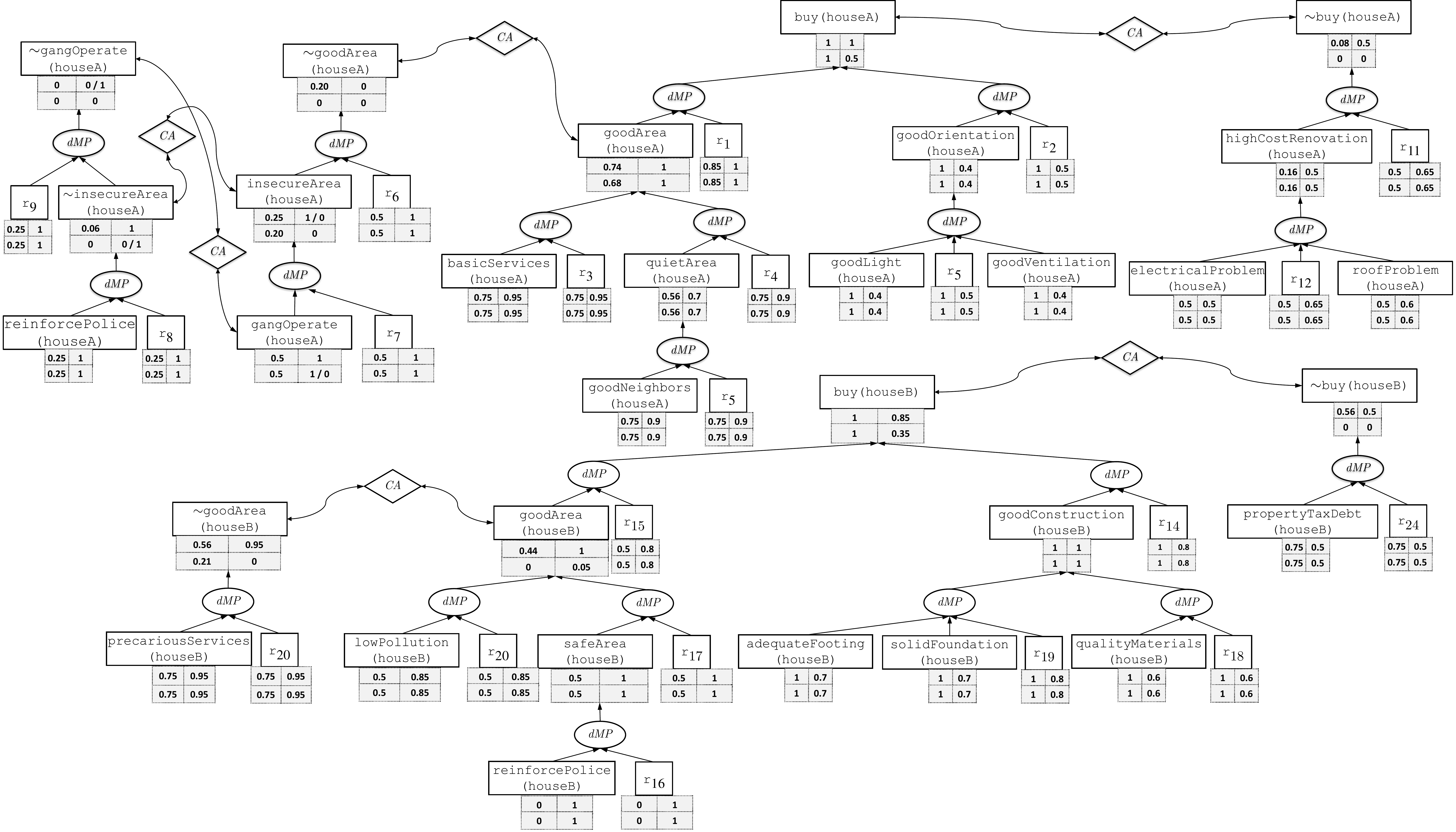}
			\caption{Representation of a labeled argumentation graph}\label{Fig.labeledargumentationgraph}
		\end{figure}
	\end{landscape}
}

\subsection{Acceptability Process in a Labeled Argumentation Framework}

The problem of understanding the process of argumentation and its role in human reasoning have been studied by many researchers in different fields including philosophy, logic, and AI~\cite{barth1982axiom,toulmin2003uses,SemAAS09}. Briefly speaking, the idea of argumentation-based reasoning is that a claim is believable if it can be defended successfully against attacking arguments. In other words, whether or not a rational agent believes in a specific claim depends on whether or not the argument supporting this claim can be successfully defended against counterarguments.

Understanding the structure and acceptability of arguments is essential for an argumentative system to be able to analyze their behavior in a specific domain. The problem of analyzing argument structure has been extensively tackled in the literature; however, the idea of introducing a formal structure to capture arguments' features---and representing how these features are affected by interactions between the arguments---is a novel one. In this way, this additional information about argument quality can be used to redefine the acceptability process, providing more information about the arguments' acceptability.

Once the I-nodes of the argumentative graph are labeled, we can consider their acceptability status based on their final valuations.
This extra information obtained through the argumentation labels is useful in analyzing an argumentation model in at least two ways: a {\em dichotomic} acceptability process and a {\em gradual} acceptability process. The former models acceptability in the classical way, with only two possible values, while the latter carries out a novel acceptability analysis modeling the acceptability as a spectrum of values. Thus, the gradual acceptability process refines the classical one.
Furthermore, we can determine the acceptability status associated with an argument taking into account the final valuation associated with the I-node that represents its conclusion, since this valuation is obtained based on the valuations of the elements that compose its structure.

Next, we introduce the formal definition of these processes in order to analyze the acceptability of the elements that comprise the argumentative graph from these two perspectives.

\begin{Definition}[Dichotomy Acceptability Status]
	\label{AcceptabilityStatusclasica}
	Let  $\Phi= \langle {\mathcal L},$ ${\mathcal R},  {\mathcal K}, {\mathcal A}, {\mathcal F} \rangle$ be an LAF,  $G_\Phi$ be the corresponding argumentation graph, and $X$ be an I-node in $G_\Phi$.
	For each of the algebras $\mathsf{A}_i$ in $\mathcal{A}$, representing a feature to be associated with each I-node $X$, let $\bot_i$ be the neutral element for the operators.
	Then, I-node $X$ has assigned one of two possible acceptability statuses according to their associated labels:
	\begin{itemize}\itemsep 3pt
		\item[--] $\mathtt{Accepted}$ if and only if $\Labelminus{i}{X} \neq \neutro_i$ for each $1 \leq i \leq n$.
		\item[--] $\mathtt{Rejected}$ if and only if $\Labelminus{i}{X} = \neutro_i$ for any $1 \leq i \leq n$.
	\end{itemize}
	We will denote with $\mathtt{S}_c$ the dichotomy status assignment to the argumentative graph $G_\Phi$.
\end{Definition}

Based on the dichotomy acceptability notion, an I-node $X$ will be $\mathtt{Accepted}$ if its final features are not neutralized by the I-node $\nega{X}$. The dichotomy acceptability status process is useful in the decision making process, since it specifies a stance on the information that the I-node represents. The following proposition is an immediate consequence of the previous definition.

\begin{Proposition}
	Let $\Phi= \LAFARG$ be an LAF, $G_\Phi$ be the corresponding argumentation graph, and \Labeling{\Phi} be a valid labeling for $G_\Phi$. Then, the dichotomy status assignment $\mathtt{S}_c$ is unique.
\end{Proposition}

\begin{Theorem}[Consistency in classical acceptability]\label{teo.consistenciaaceptadosclasica}
	Let $\Phi= \langle {\mathcal L},{\mathcal R},  {\mathcal K}, {\mathcal A}, {\mathcal F} \rangle$ be an LAF, $G_\Phi$ be the corresponding argumentation graph, $\mathtt{S}_c$ be a dichotomy status assignment for $G_\Phi$, and $X$ and $\negaBar{X}$ be two I-nodes of $G_\Phi$ representing contradictory knowledge. If $X$ is $\mathtt{Accepted}$, then $\negaBar{X}$ is $\mathtt{Rejected}$.
\end{Theorem}

\begin{Definition}[Gradual Acceptability Status]\label{AcceptabilityDegreeStatus}
	Let  $\Phi= \langle {\mathcal L},{\mathcal R},$ ${\mathcal K}, {\mathcal A}, {\mathcal F} \rangle$ be an LAF,  $G_\Phi$ be the corresponding argumentation graph, and $X$ be an I-node in $G_\Phi$. For each of the algebras $\mathsf{A}_i$ in $\mathcal{A}$, representing a feature to be associated with each I-node $X$, let $\bot$ be the neutral element for the operators.
	Then, $X$ has assigned one of four possible acceptability statuses with respect to $\mathsf{A}_i$ (in descending order of acceptability):
	
	\begin{itemize}\itemsep 3pt
		\item[--] $\mathtt{Assured}$ if and only if $\Labelminus{i}{X} = \neutros_i$.
		\item[--] $\mathtt{Unchallenged}$ if and only if $\Labelplus{i}{X} = \Labelminus{i}{X} \neq \bot_i$.
		\item[--] $\mathtt{Weakened}$ if and only if $\neutro_i < \Labelminus{i}{X} < \Labelplus{i}{X}$.
		\item[--] $\mathtt{Rejected}$ if and only if $\Labelminus{i}{X} = \neutro_i$.
	\end{itemize}
	Finally, for each claim, we form a vector with the acceptability of that claim with respect to each of the attributes, and take the least degree of those that appear in the vector as the acceptability degree for the claim as a whole. We denote with $\mathtt{S}_g$ the  gradual status assignment to the argumentative graph $G_\Phi$.
\end{Definition}

\begin{figure*}[ht!]
	\begin{center}
		\includegraphics[width=1\textwidth]{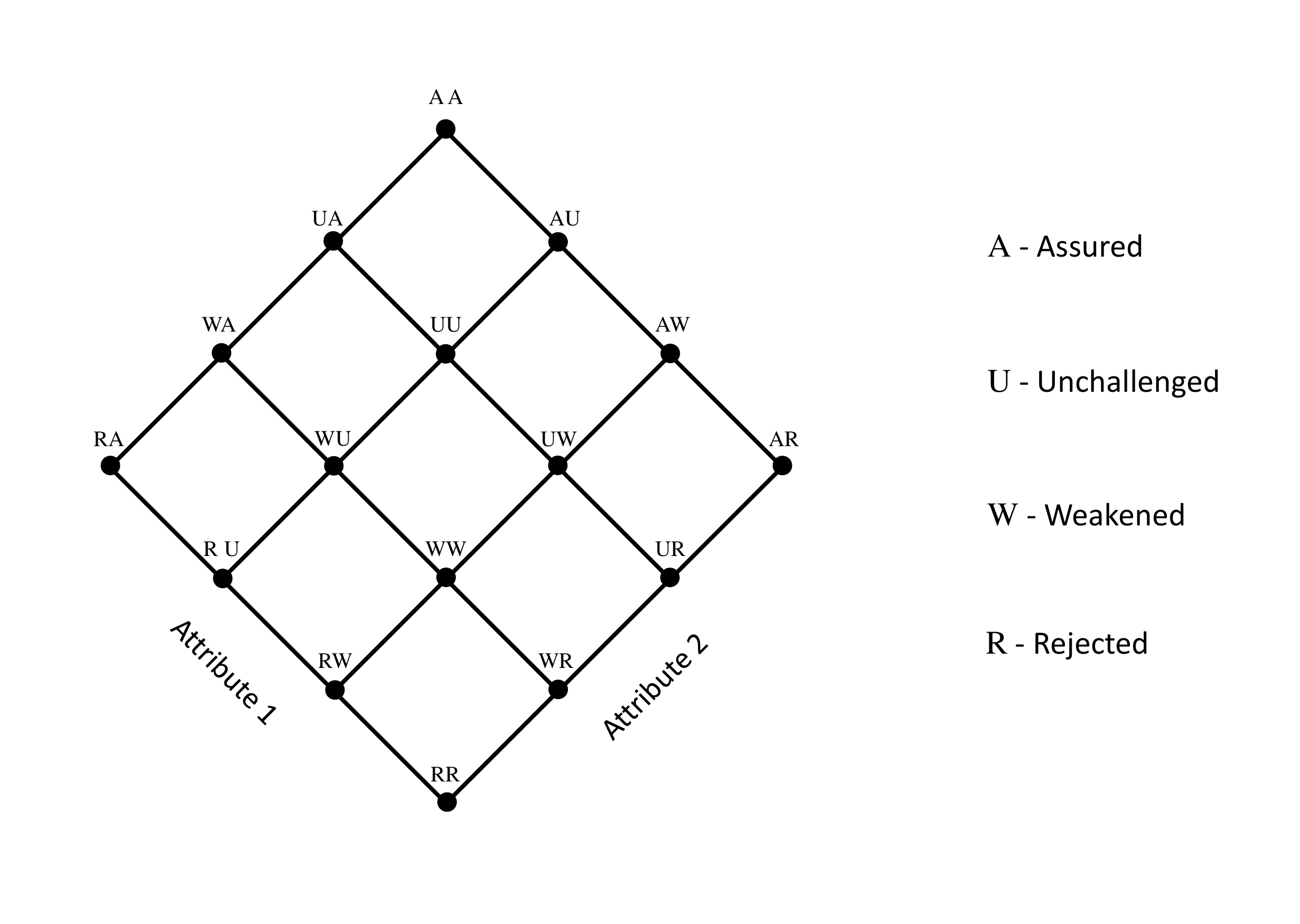}
		\caption{Status degree for a claim based on two attributes}\label{Fig.acceptabilitydegreestatus}
	\end{center}
\end{figure*}

Analyzing the gradual acceptability status process, the status of $\mathtt{Assured}$ can only be granted to an I-node $X$  when all the features associated with $X$ have a perfect valuation, while in the other end of the spectrum if $X$ has at least one feature with a neutral valuation then the claim represented through the I-node $X$ is $\mathtt{Rejected}$. In Figure~\ref{Fig.acceptabilitydegreestatus}, we illustrate Definition~\ref{AcceptabilityDegreeStatus}, where we take into account two attributes associated with a claim, in order to determine the overall status for that claim---so, each element of the lattice is a vector of length~2.

The gradual acceptability status process is useful in the decision making process because it specifies the quality of knowledge represented in a specific argumentation graph by interpreting their attributes. The following proposition is an immediate consequence of the previous definition.

\begin{Proposition}
	Let $\Phi= \LAFARG$ be an LAF, $G_\Phi$ be the corresponding argumentation graph, and \Labeling{\Phi} a valid labeling for $G_\Phi$. Then, the gradual status assignment $\mathtt{S}_g$ is unique.
\end{Proposition}

\begin{Theorem}[Consistency in gradual acceptability]\label{teo.consistenciaaceptadosgradual}
	Let $\Phi= \langle {\mathcal L},{\mathcal R},  {\mathcal K}, {\mathcal A}, {\mathcal F} \rangle$ be an LAF, $G_\Phi$ be the corresponding argumentation graph, $\mathtt{S}_g$ be a gradual status assignment for $G_\Phi$, and $X$ and $\negaBar{X}$ be two I-nodes of $G_\Phi$ representing contradictory knowledge.
	If $X$ is $\mathtt{Assured}$, $\tt{Unchallenged}$ or $\tt{Weakened}$, then $\negaBar{X}$ is $\mathtt{Rejected}$.
\end{Theorem}

\begin{Example}\label{exampleMarkedGraph}
	Let us return to Example~\ref{Ex.LabelArgumentationGraph} and calculate the acceptability status for each claim, considering a specific valid labeling for the argumentation graph that describes the argumentative discussion about buying a house. First, based on the classical acceptability status, the sets of accepted and rejected claims are:
	
	\begin{itemize}{\footnotesize
			\item[--] $S^A_c = \{\tt{gangOperate}(houseA), {\tt{reinforcePolice}(houseA)}, \tt{goodNeighbors}(houseA),\\ \tt{basicServices}(houseA), \tt{quietArea}(houseA), \tt{goodArea}(houseA), \tt{goodLight}(houseA),\\ \tt{goodVentilation}(houseA), \tt{goodOrientation}(houseA),
			{\tt{buy}(houseA)}, \tt{roofProblem}(houseA),\\ {\tt{electricalProblem}(houseA)}, \tt{highCostRenovation}(houseA),
			\tt{placeStore}(gold)\}, \\ \tt{precariousServices}(houseB), \tt{safeArea}(houseB), \tt{lowPollution}(houseB), \\ \tt{adequateFooting}(houseB), \tt{solidFoundation}(houseB), \tt{qualityMaterials}(houseB), \\ \tt{goodConstruction}(houseB), \tt{buy}(houseB), \tt{propertyTaxDebt}(houseB)	\}$
			
			\item[--] $S^R_c = \{\tt{insecureArea}(houseA), \nega{\tt{insecureArea}(houseA)}, \nega{\tt{gangOperate}(houseA)}, \\\nega{\tt{goodArea}(houseA)},
			\nega{\tt{buy}(houseA)}, \nega\tt{goodArea}(houseB), \tt{goodArea}(houseB), \\ \nega{\tt{reinforcePolice}(houseB)}, \nega{\tt{buy}(houseB)}\}$}
	\end{itemize}
	
	On the other hand, based on the acceptability degree status, the claims are classified as follows:
	
	\begin{itemize}{\footnotesize
			\item[--] $S^A_g = \{\tt{goodConstruction}(houseB)\}$
			
			\item[--] $S^U_g = \{\tt{gangOperate}(houseA), {\tt{reinforcePolice}(houseA)}, \tt{goodNeighbors}(houseA),\\ \tt{basicServices}(houseA), \tt{quietArea}(houseA),  \tt{goodLight}(houseA),\\ \tt{goodVentilation}(houseA), \tt{goodOrientation}(houseA),
			\tt{roofProblem}(houseA),\\ {\tt{electricalProblem}(houseA)}, \tt{highCostRenovation}(houseA),
			\tt{placeStore}(gold)\}, \\ \tt{precariousServices}(houseB), \tt{safeArea}(houseB), \tt{lowPollution}(houseB), \\ \tt{adequateFooting}(houseB), \tt{solidFoundation}(houseB), \tt{qualityMaterials}(houseB), \\ \tt{goodConstruction}(houseB),  \tt{propertyTaxDebt}(houseB)	\}$
			
			\item[--] $S^W_g = \{\tt{goodArea}(houseA), {\tt{buy}(houseA)}, \tt{buy}(houseB)\}$
			
			\item[--] $S^R_g = \{\tt{insecureArea}(houseA), \nega{\tt{insecureArea}(houseA)}, \nega{\tt{gangOperate}(houseA)}, \\\nega{\tt{goodArea}(houseA)},
			\nega{\tt{buy}(houseA)}, \nega\tt{goodArea}(houseB), \tt{goodArea}(houseB), \\ \nega{\tt{reinforcePolice}(houseB)}, \nega{\tt{buy}(houseB)}\}$}
	\end{itemize}
	
	Briefly speaking, the discussion presented above reflects the reasons for and against \emph{``buy house A''} and \emph{``buy house B''}. Once the acceptability status for each claim contained in the argumentation graph is determined, we analyze the status assigned to the arguments for $\tt{buy}(houseA)$ and $\tt{buy}(houseB)$ in order to specify the action that must be performed by the agent. In this sense, both are $\mathtt{Accepted}$ by the classical acceptability status process or $\mathtt{Weakened}$ by the gradual acceptability status process, justifying the investment in either house. However, we have more information about the arguments that support these decisions, which will be discussed in the next section.
\end{Example}

Note that the acceptability semantics used does not influence the composition of the set of rejected arguments.

\subsection{Modeling Context Constraints in a Labeled Argumentation Framework}

In the context of our running example, once the acceptability statuses of the arguments in favor and against ``buy house A'' or ``buy house B'' are analyzed, different scenarios may occur. If the argument in favor of buying house A is defeated by the argument against it, and the argument in favor of buying house B is strong enough to withstand its counterargument, then the agent should buy house B. In the opposite situation, when the argument in favor of buying house A is undefeated, and the argument in favor of buying house B is defeated by its opposition, then the agent should buy house A. In a third case, both arguments are defeated, so the agent should not buy either.
Finally, if both arguments are strong enough to withstand the attacks of their opponent arguments, then a conflict in the strict sense does not exist, since the arguments are not contradictory; however, the domain imposes the additional constraint that the agent only has enough money to buy one house.
To solve this, we define a preference between arguments, establishing when an argument is better than another based on their features. Next, we introduce the general notion that allows to establish a preference order (partial or total) over the arguments.

\begin{Definition}[Preference between Arguments]\label{betterargument}
	Let $\Phi= \LAFARG$ be an LAF, $G_\Phi$ be the corresponding argumentation graph, \ARGPHI\ be the set of arguments involved in $G_\Phi$, and $\mathbb{A}$ and $\mathbb{B}$ be two arguments such that $\mathbb{A}, \mathbb{B} \in \ARGPHI$, where $X$ and $Y$ are the I-nodes that represent the conclusion for the arguments $\mathbb{A}$ and $\mathbb{B}$, respectively. A function $\mathcal{P} : \ARGPHI \times \ARGPHI \rightarrow 2^{\ARGPHI}$ is a preference function that determines an order over \ARGPHI, such that:
	
	\begin{center}
		$\mathcal{P}(\mathbb{A},\mathbb{B}) = \left\{
		\begin{array}{lll}
		\{\mathbb{A}\} & \text{ iff } \Labelminus{}{X} \text{ is preferred to } \Labelminus{}{Y};\\[6pt]
		
		\{\mathbb{B}\} & \text{ iff } \Labelminus{}{Y} \text{ is preferred to } \Labelminus{}{X};\\[6pt]
		
		\{\mathbb{A},\mathbb{B}\} & \text{ iff } \Labelminus{}{X} \text{ and } \Labelminus{}{Y} \text{ are equally preferred};\\[6pt]
		
		\{\emptyset\} & \text{ iff } \Labelminus{}{X} \text{ and } \Labelminus{}{Y} \text{ are incomparable.}
		\end{array}
		\right. $
	\end{center}
\end{Definition}

Defining a preference between arguments is difficult, since in some cases their valuations are composed of more than one attribute. Different ways of solving this problem can be found in the literature; for example, analyzing the characteristics of the arguments and applying the minimax principle (MinMax and MaxMin)~\cite{yager1983entropy,kaci2008preference}, taking a strict position where an argument is preferred to another only when all its features are superior (in a sense specific)~\cite{Budan2015LAF}, among others~\cite{PrakkenSartor97,visser2012argumentation}. In this sense, the adequate comparison position should be established by the user.
Once a preference relation is established, the argumentation graph can then be completed with the corresponding PA-nodes (cf.\ Page~\pageref{page:PA-node}) so that later analyses can use this
information\label{page:preference}.

\begin{Example}
	Going back to Example~\ref{Example_InstantiateLAF}, based on the information depicted in Figure~\ref{Fig.labeledargumentationgraph} and under the condition that the agent only has enough money to buy one house, we shall determine which house the agent should buy. For this, we define a preference function $\mathcal{P}$ that reflects the agent's position regarding the information associated with the arguments establishing that an argument $\mathbb{A}$ is preferred to an argument $\mathbb{B}$ if and only if the trust degree of $\mathbb{A}$ is at least as good as the trust degree of $\mathbb{B}$ and the preference level of $\mathbb{A}$ is better than the preference level of $\mathbb{B}$ (cf.\ Table~\ref{tab.preferenceorder}).
	
	\begin{table}[ht]
		\begin{center}
			\resizebox*{!}{2.5cm}{
				\renewcommand{\arraystretch}{2}
				{\setlength{\doublerulesep}{1mm}
					\begin{tabular}{|l|c|c|}
						\hline
						\cellcolor[gray]{0.8} & \tt{buy}(houseA) & \tt{buy}(houseB)\\
						\hline \hline
						Trust Valuation & 1 & 1\\ \hline
						Preference Valuation & 0.5 & 0.35\\ \hline
			\end{tabular}}}
			\caption{Preference between Arguments}
			\label{tab.preferenceorder}
		\end{center}
	\end{table}
	
	Finally, we conclude that the accrual argument supporting the conclusion $\tt{buy}(houseA)$ is better than the argument supporting the conclusion $\tt{buy}(houseB)$, since the information exposed by the first argument is as reliable as the information provided by the second one, but at the same time the first is based on information that is more relevant to the agent, as shown in Table~\ref{tab.preferenceorder}.
\end{Example}

\subsection{Incorporating Thresholds}

As mentioned previously, in certain real-world applications agents need to make decisions to meet their goals, fulfilling requirements that depend on the domain (for instance, based on knowledge that has a certain level of trust). Towards this end, we can set a threshold that determines the minimal acceptable valuation that an argument must have to be a part of the argumentation process, verifying that each element that is part of an argument structure satisfies such threshold. The quality threshold can be incorporated in two different ways: (i) a threshold to redefine the labeling process, or (ii) a threshold to debug the acceptability process.

In the first case, an alternative version to the labeling procedure is necessary, considering a quality threshold to decide when an argument is allowed to participate in the dialectical process. A version of this procedure was already presented in~\cite{Budan2015LAF} with a simple interpretation: if the quality of some arguments obtained from the domain does not exceed a specific threshold, then such arguments do not produce any influence in the argumentative process. In this procedure, the user makes a decision about the quality of the participating arguments. In this case, the labeling procedure generates a system of equations where the quality restrictions imposed by the threshold are reflected.

In the second case, we set an acceptability threshold in order to redefine the acceptability process---i.e., once the valuations corresponding to each element involved in the dialectical process are established, we use a threshold to specify the acceptability status associated with each argument. There exist different positions to establish an acceptability threshold. For example, the valuation $\top$ represents the strongest possible valuation associated with a specific feature, but setting this as the minimum quality level could prove to be too strict even under the credulous approach. One could base the minimum quality level on the maximum or minimum of the weakened valuations of each feature, or simply take the average (of course, if this makes sense in the attribute's domain). Another possibility is to consider arguments with an intermediate valuation between $\top$ and $\bot$, capturing the intuition of majority. It is clear that each one of these positions may seem arbitrary; however, the meaning of each one depends on the application domain, the attributes associated with the arguments, and the user's interpretation. Towards this end, the dichotomy acceptability status with thresholds can be defined by adapting Definition~\ref{AcceptabilityStatusclasica} by assigning $\mathtt{Accepted}$ whenever $\Labelminus{i}{X} > \tau_i$ and $\mathtt{Rejected}$ whenever $\Labelminus{i}{X} \leq \tau_i$, where $\tau_i$ is the threshold for each attribute $1 \leq i \leq n$. Likewise, for the gradual acceptability case, we adapt Definition~\ref{AcceptabilityDegreeStatus} by assigning $\mathtt{Unchallenged}$ whenever $\Labelplus{i}{X} = \Labelminus{i}{X} \neq \tau_i$, $\mathtt{Weakened}$ whenever $\tau_i < \Labelminus{i}{X} < \Labelplus{i}{X}$, and $\mathtt{Rejected}$ whenever $\Labelminus{i}{X} = \tau_i$.

\begin{Example}
	We now return to the setting in Example~\ref{Ex.LabelArgumentationGraph}, and analyze the use case adopting the described acceptability process under the thresholds $\tau_1 = 0.7 \in A$ and $\tau_1 = 0.5 \in B$. First, based on the classical acceptability status, the sets of accepted and rejected claims are:
	
	\begin{itemize}{\footnotesize
			\item[--] $S^A_c = \{ \tt{goodNeighbors}(houseA), \tt{basicServices}(houseA), {\tt{buy}(houseA)},	\\
			\tt{placeStore}(gold)\},  \tt{precariousServices}(houseB),  \tt{adequateFooting}(houseB), \\
			\tt{solidFoundation}(houseB), \tt{qualityMaterials}(houseB), \tt{goodConstruction}(houseB), \\ \tt{propertyTaxDebt}(houseB)	\}$
			
			\item[--] $S^R_c = \{\tt{gangOperate}(houseA),{\tt{reinforcePolice}(houseA)},\tt{insecureArea}(houseA), \\
			\tt{roofProblem}(houseA), {\tt{electricalProblem}(houseA)}, \tt{buy}(houseB), \tt{quietArea}(houseA),\\
			\tt{highCostRenovation}(houseA), \tt{goodArea}(houseA), \tt{goodLight}(houseA),\\
			\tt{goodVentilation}(houseA), \nega{\tt{insecureArea}(houseA)}, \nega{\tt{gangOperate}(houseA)}, \\
			\nega{\tt{goodArea}(houseA)}, \tt{goodOrientation}(houseA), \nega{\tt{buy}(houseA)}, \\
			\nega\tt{goodArea}(houseB), \tt{goodArea}(houseB), \nega{\tt{reinforcePolice}(houseB)},\\
			\nega{\tt{buy}(houseB)}\},\tt{safeArea}(houseB), \tt{lowPollution}(houseB) \}$}
	\end{itemize}
	On the other hand, based on the acceptability degree status, the claims are classified as follows:
	\begin{itemize}{\footnotesize
			\item[--] $S^A_g = \{\tt{goodConstruction}(houseB)\}$
			
			\item[--] $S^U_g = \{ \tt{goodNeighbors}(houseA), \tt{basicServices}(houseA), \tt{precariousServices}(houseB),\\
			\tt{adequateFooting}(houseB), \tt{solidFoundation}(houseB), \tt{qualityMaterials}(houseB), \\
			\tt{propertyTaxDebt}(houseB)\}$
			
			\item[--] $S^W_g = \{{\tt{buy}(houseA)}\}$
			
			\item[--] $S^R_g = \{\tt{gangOperate}(houseA),{\tt{reinforcePolice}(houseA)},\tt{insecureArea}(houseA), \\
			\tt{roofProblem}(houseA), {\tt{electricalProblem}(houseA)}, \tt{buy}(houseB), \tt{quietArea}(houseA),\\
			\tt{highCostRenovation}(houseA), \tt{goodArea}(houseA), \tt{goodLight}(houseA),\\
			\tt{goodVentilation}(houseA), \nega{\tt{insecureArea}(houseA)}, \nega{\tt{gangOperate}(houseA)}, \\
			\nega{\tt{goodArea}(houseA)}, \tt{goodOrientation}(houseA), \nega{\tt{buy}(houseA)}, \\
			\nega\tt{goodArea}(houseB), \tt{goodArea}(houseB), \nega{\tt{reinforcePolice}(houseB)},\\
			\nega{\tt{buy}(houseB)}\},\tt{safeArea}(houseB), \tt{lowPollution}(houseB) $}
	\end{itemize}
	
	In contrast with the previous analysis, we focus on the status assigned to the arguments for and against $\tt{buy}(houseA)$ and $\tt{buy}(houseB)$ in order to specify the action that must be performed by the agent based on a quality threshold. In this sense, the argument for $\tt{buy}(houseA)$ is $\mathtt{Accepted}$ by the classical acceptability status process and $\mathtt{Weakened}$ by the gradual acceptability status process, thus signalling that there is a high-quality argument that offers relevant and trustworthy information for the user. On the other hand, the argument for $\tt{buy}(houseB)$ is $\mathtt{Rejected}$ by both the classical and gradual acceptability processes, since it has a low preference valuation, thus representing an argument that does not have valuable information for the user.
\end{Example}

In summary, the valuations associated with the arguments through labels provide us with the means to model peculiarities of an application domain. In a model, the arguments are related in different ways, and these relationships influence the information attached to them; we therefore presented a general scheme to propagate such information through the argumentation graph, allowing to obtain more details about the arguments, which in turn reveal different degrees of acceptability for them. 

\section{Discussion and Related Work}

As discussed in the introduction, the knowledge quality is an important concern in commonsense reasoning. Thus, its consideration becomes relevant when modeling argumentation capabilities of intelligent agents and systems. Next, we present some relevant work attacking the mentioned motivation, distinguishing the differences keys that exists with the formalism proposed in this work.

%\subsection{The Use of Meta-Information in the Argumentation Domain.}

D.~Gabbay's groundbreaking work on Labeled Deductive Systems~\cite{Gabbay90,Gabbay96}, has provided a clear and direct motivation for this work. The introduction of a flexible and rigorous formalism to tackle complex problems using logical frameworks that include labeled deduction capabilities has permitted to address research problems in areas such as temporal logics, database query languages, and defeasible reasoning systems. In labeled deduction, the formulas are replaced by labeled formulas, expressed as ${L\!:\!\phi}$, where $L$ represents a label associated with the logical formula $\phi$. Labels are used to carry additional information that enrich the representation language. The intuitions attached to labels may vary accordingly with the system modeling needs. The idea of structuring labels as an algebra was present from the very inception of labeled systems~\cite{Gabbay90}.

In~\cite{Gabbay96}, Gabbay's proposal was applied to argumentation systems. There, the authors proposed a framework with the main purpose of formally characterizing and comparing different argument-based inference mechanisms through a unified framework; in particular, two non-monotonic inference operators were used to model argument construction and dialectical analysis in the form of warrant. Labels were used in the framework to represent arguments and dialectical trees. Our proposal shares with those works the characteristic of also involving the use of labels together with an algebra of argumentation labels. Nevertheless, our intention is focused on pursuing a different goal; we are not trying to unify the presentation of different logics and formally compare them, but to \emph{extend} the representational capabilities of argumentation frameworks by allowing them to handle additional domain-specific information. Certainly, it can be argued that due to the extreme generality of Gabbay's framework, somehow it could also be instantiated in some way to achieve this purpose, but we have aimed in our proposal to provide a concrete framework in the context of the Argument Interchange Format, showing how to propagate labels in the specific case of the argument interactions of aggregation, support, and conflict.

In~\cite{gabbay2012numerical}, Gabbay (concerned with a different problem) proposes a numerical approach to the problem of merging of argumentation networks; he considers an augmented network containing the arguments and attacks of all the networks to be merged. Then, agents put forward their vote on the components of the network depending on how they perceive these components locally, where a vote means reinforcement in the sense that the more a component appears locally, the more it is represented globally. In addition, he presents a way to calculate the values of arguments in the weighted augmented network, and discern how the attacks to an argument affect its initial support value; finally, he presents a threshold for acceptance to determine the acceptability of an argument based on its weight. In the work presented here, we focus on arguments with structure and with the way the integration of these arguments affects acceptability; thus, our proposal shares the idea of assigning valuations to arguments, and propagating these valuations through an argumentation graph. However, in the presented framework, it is possible to associate more than one attribute to the arguments; also, we used an abstract algebraic structure in which we can perform the operations of aggregation, support, and conflict of arguments depending on the relationships that exist among them, and we determine the valuation associated with an argument through its internal structure. In addition, the conflict operator models situations in which an undefeated argument is weakened when counter-arguments exist; we used the valuation associated with arguments for determining when an argument is better than another, and we used a threshold valuation throughout the process of propagation of labels to establish the conditions that an argument must satisfy in order to be considered accepted.

In~\cite{elvang1993dialectic,cayrol2005graduality,amgoud1998acceptability,kaci2008preference,leite2011social} the authors prompted the use of a special valuations associated with the arguments to specify their strengths, and how this valuations affect the argumentative process. Next, we analyze each one of them, discriminated by their representation capability.

\subsection{Using Single Argument Valuations}

The work of~\cite{elvang1993dialectic} analyzes the fact that non-trivial arguments may be constructed for and against a specific proposition in the presence of an inconsistent database; the problem arises when determining which conclusion must be accepted. The authors define a particular concept of acceptability, which is used to reflect the different acceptability levels associated with an argument; then, they argue that ``the more acceptable an argument, the more confident we are in it''. Additionally, they define acceptability classes to assign linguistic qualifiers to the arguments. There are some similarities between this proposal and our own; starting from the consideration of a knowledge base from which it is possible to find the parts that compose an argument, the relationships that exist among the arguments are then analyzed, and the acceptability class that they belong to is determined. However, they do not take into account the domain-dependent characteristics associated with the arguments and they do not use thresholds to further inform the argumentative process. In another related work~\cite{krause1995logic} the authors propose a formalism in which ``arguments have the form of logical proof, but they do not have the force of logical proof ''; thus, they present a concrete formal model for practical reasoning in which a structured argument---rather than some measure---is used for describing uncertainty, \ie the degree of confidence in a proposition is obtained by analyzing the structure of the arguments relevant to it. In this way, their formalism is focused on the representation of uncertainty, and proposes a way to  calculate the aggregation of reasons for a certain proposition. In our formalism, depending on the application domain, we introduce a general framework in which it is possible to instantiate each of the elements to represent various attributes associated with the arguments; in addition, the formalism not only allows the aggregation of the valuations associated with a particular conclusion, but also provides ways to affect the weakening of the valuations for a conclusion produced by the existence of reasons against it.

In~\cite{cayrol2005graduality}, a two-step argumentation process is described: (i) the calculation of a valuation of the relative strength of the arguments, and (ii) the selection of the most acceptable among them. The focus is on defining a gradual valuation of arguments based on their interactions, and then establishing a graded concept of acceptability of arguments. The authors assert that an argument is all the more acceptable if it can be preferred to its attackers, and propose a domain of argument valuations where aggregation and reduction operators are defined; however, they do not consider the argument structure, and the evaluation of the arguments are solely based on their interaction. In our work, we determine the valuation of arguments through their internal structure, considering the different interactions among them, and propagating the valuations associated with the arguments using the operations defined in the algebra of argumentation labels. It is important to note that, unlike the proposal of Cayrol and Lagasquie-Schiex, the operations assigned to each relation among arguments are defined by the user---this provides the possibility of explicitly considering the domain of the problem. Moreover, we provide the ability of assigning more than one valuation to the arguments, depending on the features we wish to model. In addition, our formalism give the possibility to obtain a feasible solution that minimizes (or maximizes) some measure representing the different point of view of a particular user. Finally, after analyzing all the interactions among arguments we obtain final valuations assigned to each argument; then, through these valuations the acceptability status (\emph{assured, unchallenged, weakened,} or \emph{rejected}) of the arguments is obtained.

The ability of the abstract argumentative framework proposed by Dung to analyze and treat the inconsistency associated with a knowledge base is well-known; this framework identifies the arguments that describe a specific argumentation discussion, and establishes a defeat relation between conflicting arguments. Then, through a collective semantic process, it analyzes the discussion in order to determine the acceptability status (accepted or rejected) associated with the involved arguments. However, due to the high-level abstraction taken by this formalism, it is not possible to analyze the arguments' acceptability from an individual perspective since their individual properties unknown. L.~Amgoud and C.~Cayrol proposed a formalism, called \emph{Preference-Based Argumentation Framework}~\cite{amgoud1998acceptability}, where a preference relation between the arguments is introduced in the classical abstract argumentation framework in order to consider the user's preferences. The defeat relation represents a conflict based on purely logical properties (such as ``rebut'' or ``undercut'', for instance), while the preference relation represents the preferences (meta-knowledge) that cannot be extracted from the arguments themselves. Then, using a combination of these relations, it is possible to conclude that an argument defends itself when it is preferred over all its attackers, or an argument is defended when there exists a set of arguments that defeats its attackers, being the arguments of the defending set preferred to the arguments of the attacking set. In contrast, our formalism details the arguments' features that are used as a tool to specify preferences between arguments with the purpose of satisfying a specific set of contextual constraint. Note that in our formalism the preference function is not used in the conflict resolution process, since this role is performed by the conflict operator defined in the algebra of argumentation labels, which specifies how the valuations of the arguments involved in a conflict relation are affected.

T.~J.~M.~Bench-Capon and J.~L.~Pollock have introduced systems that currently have great influence over research in argumentation; we will discuss them in turn. Bench-Capon~\cite{VBAF-BC} argues in his research that oftentimes it is impossible to conclusively demonstrate in the context of disagreement that either party is wrong, particularly in situations involving practical reasoning. The fundamental role of argument in such cases is to persuade rather than to prove, demonstrate, or refute. In his own words:
%\begin{quote}
``{\em The point is that in many contexts the soundness of an argument is not the only consideration: arguments also have a force which derives from the value they advance or protect}.''
%\end{quote}
Based on this intuition, the authors propose a formalism, called \emph{Valued-Based Argumentation Framework} (\emph{VAF}), extending Dung's model to consider the strength of arguments and, through these assessments, reflect the preference of the audience to which the arguments are directed. Specifically, in \emph{VAF}, an argument has associated a value from some set that is has an ordering based on a specific audience. Then, from the valuations assigned to the arguments and the preferences defined by the audience, it is possible to specify when an argument is strong enough to attack and defeat another argument. Therefore, different audiences specify different orders over the set of values, determining different defeat relations between arguments. Finally, the authors also propose a set of semantics that extend the classical ones, defining two kinds of acceptable sets of arguments: one considering those arguments accepted by all the audiences (arguments accepted objectively), and another containing those arguments that are accepted by at least one audience (arguments accepted subjectively). In contrast, our formalism presents the tools necessary to associate a collection of valuations with the arguments representing their special features, reaching a generalization of the concept of value introduced in \emph{VAF}. Moreover, each feature is represented and computed by an algebra of labels that determines how this feature is affected by the arguments' interaction---in particular, \emph{LAF} analyzes the interactions between arguments modeling their strengthening and weakening in the domain. Finally, we use the information expressed in the labels to establish different acceptability degrees or improve the discussion via a pruning process that considers only the most relevant arguments.

\subsection{Using Multiple Argument Valuations}

In~\cite{kaci2008preference}, S.~Kaci and L.~van der Torre generalize Bench-Capon's value-based argumentation theory such that arguments can promote multiple values, and preferences among values or arguments can be specified in various ways. Each value can be associated with one or more arguments and viceversa. Then, once the different values are mapped to each argument involved in the discussion, the existing conflict relations are analyzed with the intention of identifying the successful attacks. In Bench-Capon's value-based framework, the attack of an argument $\mathbb{A}$ over an argument $\mathbb{B}$ is successful if and only if $\mathbb{A}$ attacks $\mathbb{B}$ and the value promoted by $\mathbb{B}$ is not preferred over that promoted by $\mathbb{A}$. However, in this new proposal, the arguments can promote more than one value---this increases the difficulty of determining when an argument is preferred to another based on their valuations. To address this problem the authors provide two guidelines, based on the principles of minimal/maximal specificity, that allow to establish a unique possible ordering (total ordering) over the set of values associated with the arguments. Then, based on this order, it is possible to obtain the successful attacks and subsequently the acceptable	arguments; this is done by combining algorithms from non-monotonic reasoning with others for calculating extensions in abstract argumentation. The main idea of the authors has certain similarities with our work: the valuations associated with the arguments provide the possibility of establishing argument strength, and the arguments have associated different valuations representing attributes that not are related with the logical soundness of the arguments. However, we think that each of the features associated with the arguments must have a particular interpretation, an own ordering, and an individual treatment. In this sense, a set of algebras of argumentation labels is provided to represent and compute the different characteristics of the arguments modeling the knowledge behavior in the argumentation domain.

%In ~\cite{Pollock2010}, Pollock points out the fact that, in defeasible reasoning, most semantics ignore the issue of the inner force of arguments, i.e., that some arguments support their conclusions more strongly. But once we acknowledge that arguments can differ in strength and conclusions can differ in their degree of justification, things become more complicated. In particular, he introduces the notion of diminishers, which are defeaters that cannot completely defeat their target, but instead lower the degree of justification of that argument''.

Motivated by the idea of encouraging and enhancing a debate on a particular topic in social networks, J.\ Leite and J.\ Martins proposed in~\cite{leite2011social} an extension of Dung's framework with the possibility to associate votes to arguments, together with a semantics that assigns a value to each argument; such values are drawn from a predetermined set of possible values, and represent the arguments strength (taking into account both the structure of the graph and the social opinion expressed through the votes). This proposal has some similarities with the one presented in this paper, since both obtain additional information regarding the quality of the arguments; however, LAF can be interpreted as a generalization of the social argumentative framework allowing to represent the multiple features associated with the arguments, such as use preferences and the accuracy of the information that the arguments represent, among others. Furthermore, some similarities can be found between the operations used to manipulate the social strength of the arguments and the operations defined in the algebra of argumentation labels; nevertheless, conflict resolution is modeled in a different way. In~\cite{leite2011social}, the authors model a depreciation of the social values associated with conflicting arguments without establishing a convincing defeat, while our formalism provides a rational and conventional resolution where the attributes associated with the conflicting arguments establish a convincing defeat modeling a weakening of the winning argument and a neutralization of the defeated argument.

\subsection{Fuzzy and Probabilistic Frameworks}

There have been many developments centered around the extension of argumentation frameworks---both structured and abstract---with machinery for representing and reasoning with fuzzy and probabilistic information.

Works that discuss the use of fuzzy logic can be seen in two groups: those that use fuzzy sets and relations to refine Dung's semantics (\cite{janssen2008fuzzy},\cite{tamani2014fuzzy}), and those that use fuzzy logic to assign weights to different parts of the reasoning mechanism behind the construction of arguments (\cite{Schroeder01fuzzyargumentation,Schweimeier04fuzzy}, \cite{2004uai,AlsinetCGSS08-2}), as we do here. The main innovations with respect to the latter are that we allow for a flexible use of different operations and introduce the conflict operation to weaken claims in the knowledge base.

In a similar fashion, probabilistic argumentation frameworks can be divided into those that arise from extending abstract models~\cite{Thimm12,Hunter12,FazzingaFP15} and others based on incorporating probabilities in structured frameworks---the earliest such works appeared almost two decades ago~\cite{krause:ci,haenni1999probarg}, followed by a period of inactivity and a recent resurgence of interest in the topic~\cite{Hunter13,ShakarianSMP15,SimariSF16}.
Clearly, our work is closest in spirit to the latter, since labels can be seen as a generalization of probability values associated with items in the knowledge base---the generalization is not, however, a complete subsumption since the algebra in this case would simply model a probabilistic space; the modeling of the corresponding probability distribution needs to reside elsewhere, as done in the works mentioned above.

\bigskip

Considering the intuitions of all these research lines, we formalized the foundations for an argumentative framework that integrates AIF into the system. Labels provide a way of representing salient characteristics of the arguments, generalizing the notion of value. Using this framework, we are able to establish argument acceptability, where the final labels propagated to the accepted arguments provide additional acceptability information, such as degree of justification, restrictions on justification, and others. 

\section{Conclusions}\label{Conclusion_work}

In this paper, we proposed the idea that giving an argumentation formalism more representational capabilities can enhance its use in different applications that require different elements to support conclusions. For instance, in an agent implementation, it would be beneficial to establish a measure of the success obtained by reaching a given objective; or, in the domain of recommender systems, it is interesting to provide recommendations together with an uncertainty measure or a reliability degree associated with them.

Towards realizing this idea, our work has focused on the development of a \emph{Labeled Argumentation Framework} (LAF), combining the knowledge representation capabilities provided by the \emph{Argument Interchange Format} (\emph{AIF}) together with the management of labels by an algebra developed for that purpose. We have associated operations in an algebra of argumentation labels to three different types of argument interactions, allowing to propagate information in the argumentation graph. From the algorithm used to label an argumentation graph, it is possible to determine the acceptability of arguments and the resulting extra data associated with them. A peculiarity of the conflict operation defined in the algebra is that it allows the weakening of arguments, which contributes to a better representation of application domains.

In our running example we used the reliability of the source and a preference measure associated with the arguments supporting decision making. We defined when an argument is better than another based on these features, and we considered the use of a quality threshold in two ways: (i) for deciding when an argument is involved in the argumentative discussion, and (2) for redefining the acceptability process.

Finally, we are currently developing an implementation of LAF extending the existing DeLP system~\cite{GarciaSimari2014}, which is the basis of our work\footnote{See http://lidia.cs.uns.edu.ar/delp}. The resulting implementation will be exercised in different domains that require modeling extra information associated with the arguments, taking as motivation the studies and analyses of P-DeLP~\cite{Chesnevar2004,AlsinetCGS08}, and DeLP3E~\cite{Shak:AMAI:2016}---DeLP has already been shown to have practical application in the cyber security somain~\cite{Shak:AICS:2016}. 

\bigskip
\noindent
{\bf Acknowledgments.}
This work has been partially supported by EU H2020 research and innovation programme under the Marie Sklodowska-Curie grant agreement No.\ 690974 for the project MIREL: MIning and REasoning with Legal texts, and by funds provided by CONICET, Universidad Nacional del Sur and Universidad Nacional de Santiago del Estero, Argentina.
Some of the authors of this work were also supported by the U.S.\ Department of the Navy, Office of Naval Research,
grant N00014-15-1-2742.
Any opinions, findings, and conclusions or recommendations expressed in this material are those of the
authors and do not necessarily reflect the views of the Office of Naval Research.

\bibliographystyle{elsarticle-num}
\bibliography{bib}

\appendix
\section{Proofs}
\setcounter{Proposition}{0}
\setcounter{Lemma}{0}
\setcounter{Theorem}{0}

\begin{Proposition}
	The worst-case running time of Algorithm~1 is $O(m\times t)$, where $m$ is the number of I-nodes and $t$ is the
	number of RA-nodes in the graph.
\end{Proposition}

\noindent\underline{Proof}: Intuitively, as a loose upper bound, we can say that the labeling process for an argumentation graph $G$ has a worst-case running time in $O(n^3)$. This arises from the following analysis: first, we need to label each I-node ($O(n)$); second, for each I-node we must analyze all the RA-nodes associated with it ($O(n)$); third, for each RA-node we must analyze all the premises associated with it ($O(n)$); and fourth, for contradictory literals we analyze and obtain the corresponding weakness value ($O(1)$).

However, consider the {\em entire} set of nodes (I, RA, and CA); let $n$ denote the cardinality of this set, with $m$ I-nodes, $t$ RA-nodes, and $k$ CA-nodes. We can thus refine the above $O(n)$ terms to $O(m)$, $O(t)$, and $O(m)$, respectively, and the computational complexity of the labeling process is more accurately described as $O(m^2 \times t)$. Note that when modeling real-world examples there is usually an upper bound on the number of premises that support an I-node through the application of an inference rule represented by an RA-node. Note that the number of premises that support an I-node through the application of an inference rule cannot be arbitrarily large---this upper bound can be defined as $p=5$ in the worst case, so for the purpose of this analysis we can assume $p \in O(1)$. Therefore, we can conclude that the computational complexity of the labeling process is $O(m\times t)$. $\hfill\Box$

\begin{Theorem}[Convergence]
	Let $\Phi= \LAFARG$ be an LAF, $G_\Phi$ be the corresponding argumentation graph for $\Phi$, and $S_1$ and $S_2$ be two different labeling sequences for $G_\Phi$, where $S_1$ generates system $EQS_1$ and $S_2$ generates system $EQS_2$.
	Then, we have that $EQS_1 = EQS_2$.
\end{Theorem}

\noindent\underline{Proof}: Let us suppose that $EQS_1 \neq EQS_2$. Then, either $EQS_1 \nsubseteq EQS_2$, or $EQS_2 \nsubseteq EQS_1$, or both; assume, without loss of generality, that $EQS_1 \nsubseteq EQS_2$. By Definition~\ref{Def.Labelgraphcycle}, the labeling procedure does not depend on the order in which the argumentation graph is analyzed, but this procedure depends on the relation $($support, conflict, and aggregation$)$ between the elements that compose the knowledge base \K\ or derived from \K\ through \R. In this way, if there exists an equation in $EQS_1$ that is not in $EQS_2$, then there exists a relation analyzed by sequence $S_1$ that was not analyzed by sequence $S_2$---this is a contradiction, since $S_1$ and $S_2$ are sequences obtained by a labeling process that explores the entire argumentative graph $G_\Phi$ (in depth and breadth).
$\hfill\Box$

\begin{Lemma}
	Let $\Phi= \LAFARG$ be an LAF, $G_\Phi$ be the corresponding argumentation graph for $\Phi$, \model\ be a valid labeling for $G_{\Phi}$, and $X$ be an I-node in $G_{\Phi}$. Then, the labels \Labelplus{i}{X} and \Labelminus{i}{X} representing each feature of $X$ through an algebra $\mathsf{A}_i$ in $\mathcal{A}$ satisfy the following properties:
	\begin{itemize}
		\item[i)]  $\Labelminus{i}{X} \leq \Labelplus{i}{X}$ (thus, if $\Labelplus{i}{X}=\bot$, it follows that $\Labelminus{i}{X} =\bot$ as well).
		\item[ii)]  If $\Labelplus{i}{\overline{X}} = \IE$, then $\Labelminus{i}{X} = \Labelplus{i}{X}$.
		\item[iii)]  If $\Labelplus{i}{X} \leq \Labelplus{i}{\overline{X}}$, then $\Labelminus{i}{X} =\bot$.
	\end{itemize}
\end{Lemma}

\noindent
\underline{Proof}:
Recall that an I-node $X$ has two associated valuations, \Labelplus{i}{X} and \Labelminus{i}{X}, where \Labelplus{i}{X} represents an accumulation valuation specifying the aggregation of the reasons supporting the claim $X$, while \Labelminus{i}{X} displays the state of the claim $X$ after considering the conflict resolution. Thus, the labeling procedure introduced in Definition~\ref{Def.Labelgraphcycle} establishes a dependency between \Labelplus{i}{X} and \Labelminus{i}{X}, since to be able to compute the weakened valuation for the claim $X$ it is necessary to know the accumulated valuations for $X$ and $\overline{X}$. Then, by the conditions that a conflict operator must satisfy introduced in Definition~\ref{Def.Algebra_Labels}, we obtain that $\Labelplus{i}{X} \ \conflicto \ \Labelplus{i}{\overline{X}} \le \Labelplus{i}{X}$, and thus property (i) holds.

By (i), we know that \Labelminus{i}{X} $\leq$ \Labelplus{i}{X}. In addition, by Definition~\ref{Def.Algebra_Labels} the domain of labels is composed of the top element $\top_i$ representing the maximum label with respect to a partial order $\le_i$ and the neutral element for the operator \support, while $\bot_i$ is the minimum label and the neutral element for the operators \conflicto\ and \agregation. Then, if $\ \Labelplus{i}{X} = \bot_i\ $  and  $\ \Labelplus{i}{X} \ \conflicto \ \Labelplus{i}{\overline{X}} \le \Labelplus{i}{X}$ then \Labelminus{i}{X}$= \neutro_i$.
Also, if $\ \Labelplus{i}{\overline{X}} = \bot_i\ $  then  $\ \Labelplus{i}{X} \ \conflicto \ \Labelplus{i}{\overline{X}} = \Labelplus{i}{X}$, since $\bot_i$ is the neutral element for the operator \conflicto, satisfying property (ii).

Finally, based on the conditions defined for the construction of the conflict operator \conflicto, we have that: if the accrued value of $X$ is at most as strong as the accrued value of $\overline{X}$ $($with respect to $\le_i$$)$, then the strength of $X$ is neutralized by the strength of $\overline{X}$, satisfying property (iii).
$\hfill\Box$

\begin{Theorem}[Consistency in a classical acceptability procedure]
	Let \linebreak $\Phi=\langle {\mathcal L},{\mathcal R},  {\mathcal K}, {\mathcal A}, {\mathcal F} \rangle$ be an LAF, $G_\Phi$ be the corresponding argumentation graph, $\mathtt{S}_c$ be a dichotomy status assignment for $G_\Phi$, and $X$ and $\negaBar{X}$ be two I-nodes of $G_\Phi$ representing contradictory knowledge. If $X$ is $\mathtt{Accepted}$, then $\negaBar{X}$ is $\mathtt{Rejected}$.
\end{Theorem}

\noindent
\underline{Proof}:
Let us suppose that $X$ and \nega{X} both have the $\mathtt{Accepted}$ status. Then, by Definition~\ref{AcceptabilityStatusclasica}, for all attributes $i$ it holds that $\Labelminus{i}{X} \neq \neutro_i$ and $\Labelminus{i}{\negaBar{X}} \neq \neutro_i$.

\begin{itemize}\itemsep 3pt
	\item[--] First, suppose that $\Labelminus{i}{X} \leq \Labelminus{i}{\negaBar{X}}$. By Definition~\ref{Def.Labelgraphcycle}, we have that $\Labelminus{i}{X} = \Labelplus{i}{X} \conflicto \Labelplus{i}{\negaBar{X}}$ and $\Labelminus{i}{\negaBar{X}} = \Labelplus{i}{\negaBar{X}} \conflicto \Labelplus{i}{X}$. We can deduce that $\Labelplus{i}{X} \conflicto \Labelplus{i}{\negaBar{X}} \leq  \Labelplus{i}{\negaBar{X}} \conflicto \Labelplus{i}{X}$, but this occurs if $\Labelplus{i}{X} \leq \Labelplus{i}{\negaBar{X}}$. Hence, by property (iv) of Lemma~\ref{prop.values} we have that $\Labelminus{i}{X} = \bot_i$. Finally, by Definition~\ref{AcceptabilityStatusclasica}, $X$ has the status $\mathtt{Rejected}$ and $\negaBar{X}$ has the status $\mathtt{Accepted}$, which is a contradiction.
	
	\item[--] Second, suppose that $\Labelminus{i}{X} > \Labelminus{i}{\negaBar{X}}$. By Definition~\ref{Def.Labelgraphcycle}, we have that $\Labelminus{i}{X} = \Labelplus{i}{X} \conflicto \Labelplus{i}{\negaBar{X}}$ and $\Labelminus{i}{\negaBar{X}} = \Labelplus{i}{\negaBar{X}} \conflicto \Labelplus{i}{X}$. We can then deduce that $\Labelplus{i}{X} \conflicto \Labelplus{i}{\negaBar{X}} >  \Labelplus{i}{\negaBar{X}} \conflicto \Labelplus{i}{X}$, but this occurs if $\Labelplus{i}{X} > \Labelplus{i}{\negaBar{X}}$. Hence, by property (iv) of Lemma~\ref{prop.values} we have that $\Labelminus{i}{\negaBar{X}} = \bot_i$. Finally, by Definition~\ref{AcceptabilityStatusclasica}, $\negaBar{X}$ has the status $\mathtt{Rejected}$ and $X$ has the status $\mathtt{Accepted}$, which is a contradiction.
\end{itemize}		
We can thus conclude that two I-nodes representing contradictory information cannot both have the $\tt{Accepted}$ status.
$\hfill\Box$

\begin{Theorem}[Consistency in a gradual acceptability procedure]
	Let \linebreak
	$\Phi= \langle {\mathcal L},{\mathcal R},  {\mathcal K}, {\mathcal A}, {\mathcal F} \rangle$ be an LAF, $G_\Phi$ be the corresponding argumentation graph, $\mathtt{S}_g$ be a gradual status assignment for $G_\Phi$, and $X$ and $\negaBar{X}$ be two I-nodes of $G_\Phi$ representing contradictory knowledge.
	If $X$ is $\mathtt{Assured}$, $\tt{Unchallenged}$ or $\tt{Weakened}$, then $\negaBar{X}$ is $\mathtt{Rejected}$.
\end{Theorem}

\noindent
\underline{Proof}:
We need to show that two I-nodes that represent contradictory formulas cannot both be accepted (to any degree). We divide the proof into the following cases:

\begin{itemize}\itemsep 3pt
	\item[--] Suppose that $X$ and $\negaBar{X}$ have status $\mathtt{Assured}$ w.r.t.\ attribute $i$. By hypothesis, we have that $\Labelminus{i}{X} = \top_i$ and $\Labelminus{i}{\negaBar{X}} = \top_i$. Also, by item (i) of Lemma~\ref{prop.values}, we establish that $\Labelplus{i}{X} \geq \Labelminus{i}{X}$, while by Definition~\ref{Def.Labelgraphcycle} we calculate the weakened values for the I-nodes $X$ and $\negaBar{X}$ through the equations $\Labelminus{i}{X} = \Labelplus{i}{X} \ \conflicto \ \Labelplus{i}{\negaBar{X}}$ and $\Labelminus{i}{\negaBar{X}} = \Labelplus{i}{\negaBar{X}} \ \conflicto \ \Labelplus{i}{X}$. Then, we can deduce that $\Labelminus{i}{X} = \top_i \conflicto \top_i$ and $\Labelminus{i}{\negaBar{X}} = \top_i \conflicto \top_i$. Finally,  by the conditions that a conflict operator must satisfy (Definition~\ref{Def.Algebra_Labels}) we can conclude that $\Labelminus{i}{X} = \bot_i$ and $\Labelminus{i}{\negaBar{X}} = \bot_i$, which is a contradiction.
	
	\item[--] Suppose that $X$ has status $\mathtt{Assured}$ and $\negaBar{X}$ has status $\mathtt{Unchallenged}$. By hypothesis, we have that $\Labelminus{i}{X} = \top_i$ and $\Labelminus{i}{\negaBar{X}} = \Labelplus{i}{\negaBar{X}} \neq \bot_i$. Also, by item (i) of Lemma\ref{prop.values}, we can establish that $\Labelplus{i}{X} \geq \Labelminus{i}{X}$, while by Definition~\ref{Def.Labelgraphcycle} we can calculate the weakened values for the I-nodes $X$ and $\negaBar{X}$ through the equations $\Labelminus{i}{X} = \Labelplus{i}{X} \ \conflicto \ \Labelplus{i}{\negaBar{X}}$ and $\Labelminus{i}{\negaBar{X}} = \Labelplus{i}{\negaBar{X}} \ \conflicto \ \Labelplus{i}{X}$. Then, we can deduce that $\Labelminus{i}{X} = \top_i \ \conflicto \ \Labelplus{i}{\negaBar{X}}$ and $\Labelminus{i}{\negaBar{X}} = \Labelplus{i}{\negaBar{X}} \ \conflicto \ \top_i$. Finally,  by the conditions that a conflict operator must satisfy, we can conclude that $\Labelminus{i}{X} \leq \Labelplus{i}{X}$ and $\Labelminus{i}{\negaBar{X}} = \bot_i$, leading to a contradiction.
	
	\item[--] Suppose that $X$ has status $\mathtt{Assured}$ and $\negaBar{X}$ has status $\mathtt{Weakened}$. By hypothesis, we have that $\Labelminus{i}{X} = \top_i$ and $\bot_i < \Labelminus{i}{\negaBar{X}} < \Labelplus{i}{\negaBar{X}}$. Also, by item (i) of Lemma~\ref{prop.values}, we can establish that $\Labelplus{i}{X} \geq \Labelminus{i}{X}$, while by Definition~\ref{Def.Labelgraphcycle} we can calculate the weakened values for the I-nodes $X$ and $\negaBar{X}$ through the equations $\Labelminus{i}{X} = \Labelplus{i}{X} \ \conflicto \ \Labelplus{i}{\negaBar{X}}$ and $\Labelminus{i}{\negaBar{X}} = \Labelplus{i}{\negaBar{X}} \ \conflicto \ \Labelplus{i}{X}$. Then, we can deduce that $\Labelminus{i}{X} = \top_i \ \conflicto \ \Labelplus{i}{\negaBar{X}}$ and $\Labelminus{i}{\negaBar{X}} = \Labelplus{i}{\negaBar{X}} \ \conflicto \ \top_i$. Finally,  by the conditions that a conflict operator must satisfy, we can conclude that $\Labelminus{i}{X} \leq \Labelplus{i}{X}$ and $\Labelminus{i}{\negaBar{X}} = \bot_i$, which is a contradiction.
	
	\item[--] Suppose now that $X$ and $\negaBar{X}$ both have status $\mathtt{Unchallenged}$.  By hypothesis, we have that $\Labelminus{i}{X} = \Labelplus{i}{X} \neq \bot_i$ and $\Labelminus{i}{\negaBar{X}} = \Labelplus{i}{\negaBar{X}} \neq \bot_i$. Also, by item (i) of Lemma~\ref{prop.values} we can establish that $\Labelminus{i}{X} = \Labelplus{i}{X} \ \conflicto \ \Labelplus{i}{\negaBar{X}}$ and $\Labelminus{i}{\negaBar{X}} = \Labelplus{i}{\negaBar{X}} \ \conflicto \ \Labelplus{i}{X}$. Finally,  using the conditions that a conflict operator must satisfy, we can conclude that $\Labelminus{i}{X} < \Labelplus{i}{X}$ and $\Labelminus{i}{\negaBar{X}} < \Labelplus{i}{\negaBar{X}}$, and thus a contradiction.
	
	\item[--] Suppose that $X$ has status $\mathtt{Unchallenged}$ and $\negaBar{X}$ has status $\mathtt{Weakened}$. By hypothesis, we have that $\Labelminus{i}{X} = \Labelplus{i}{X} \neq \bot_i$ and $\bot_i < \Labelminus{i}{\negaBar{X}} < \Labelplus{i}{\negaBar{X}}$. Moreover, by item (i) of Lemma~\ref{prop.values}, we are able to establish that $\Labelminus{i}{X} = \Labelplus{i}{X} \ \conflicto \ \Labelplus{i}{\negaBar{X}}$ and $\Labelminus{i}{\negaBar{X}} = \Labelplus{i}{\negaBar{X}} \ \conflicto \ \Labelplus{i}{X}$. Finally,  by the conditions that a conflict operator must satisfy, we can conclude that $\Labelminus{i}{X} < \Labelplus{i}{X}$ and $\Labelminus{i}{\negaBar{X}} < \Labelplus{i}{\negaBar{X}}$---a contradiction.
	
	\item[--] Suppose that $X$ and $\negaBar{X}$ both have acceptability status $\mathtt{Weakened}$. By hypothesis, we have that $\bot_i < \Labelminus{i}{X} < \Labelplus{i}{X}$ and $\bot_i < \Labelminus{i}{\negaBar{X}} < \Labelplus{i}{\negaBar{X}}$. Also, by item (i) of Lemma~\ref{prop.values} we can establish that $\Labelminus{i}{X} = \Labelplus{i}{X} \ \conflicto \ \Labelplus{i}{\negaBar{X}}$ and $\Labelminus{i}{\negaBar{X}} = \Labelplus{i}{\negaBar{X}} \ \conflicto \ \Labelplus{i}{X}$. We must analyze two possible cases:
	
	\begin{itemize}\itemsep 2pt
		\item If $\Labelplus{i}{X} < \Labelplus{i}{\negaBar{X}}$, from the conditions that a conflict operator must satisfy, we can conclude that $\Labelminus{i}{X} = \bot_i$ and $ \bot_i < \Labelminus{i}{\negaBar{X}} < \Labelplus{i}{\negaBar{X}}$, which is a contradiction.
		
		\item Now suppose that $\Labelplus{i}{X} \geq \Labelplus{i}{\negaBar{X}}$. By the conditions that a conflict operator must satisfy, we can conclude that $\Labelminus{i}{X} = \bot_i$ y $ \Labelminus{i}{\negaBar{X}} = \bot_i$ \'o $\Labelminus{i}{X} < \Labelplus{i}{X}$ y $ \Labelminus{i}{\negaBar{X}} = \bot_i$, a contradiction.
	\end{itemize}
\end{itemize}
These contradictions arise from assuming that two I-nodes representing contradictory information can be accepted simultaneously under any feature.

Finally, by Definition~\ref{AcceptabilityDegreeStatus}, the final acceptability status for an I-node is obtained by applying a skeptic policy under its acceptability status vector. Suppose that $X$ and $\negaBar{X}$ have acceptability status $\tt{Assured}$. Then, $\tt{Assured}$ is the lowest acceptability degree of all I-nodes involved. However, by the above points, if the lowest acceptability degree for I-node $X$ is $\tt{Assured}$, then the status assigned to I-node $\negaBar{X}$ must be $\tt{Rejected}$,
being this the lowest acceptability degree for the I-node $\negaBar{X}$, which is a contradiction. This contradiction arises from assuming that two I-nodes representing contradictory information can be accepted simultaneously.
$\hfill\Box$ 

\end{document}